\documentclass{article}

\usepackage{microtype}
\usepackage{graphicx}
\usepackage{subcaption}

\usepackage{hyperref}

\usepackage[preprint]{icml2026}

\usepackage{amsmath}
\usepackage{amssymb}
\usepackage{mathtools}
\usepackage{amsthm}

\DeclareMathOperator*{\argmax}{arg\,max}
\DeclareMathOperator*{\argmin}{arg\,min}

\usepackage[utf8]{inputenc} % allow utf-8 input
\usepackage{hyperref}       % hyperlinks
\usepackage{url}            % simple URL typesetting
\usepackage{booktabs}       % professional-quality tables
\usepackage{amsfonts}       % blackboard math symbols
\usepackage{nicefrac}       % compact symbols for 1/2, etc.
\usepackage{microtype}      % microtypography
\usepackage{xcolor}         % colors
\usepackage{amsmath,amssymb,amsthm}
\usepackage{booktabs}
\usepackage{pifont}
\usepackage{xcolor}
\usepackage{booktabs} % for professional tables
\usepackage{array}
\usepackage{multirow}
\usepackage{enumitem}
\usepackage{changepage}

\newcommand{\AlgName}{\textsc{LOCUS}}

\usepackage[capitalize,noabbrev]{cleveref}

\theoremstyle{plain}

\theoremstyle{definition}

\theoremstyle{remark}

\usepackage[textsize=tiny]{todonotes}

\crefname{equation}{eq.}{eqs.}
\Crefname{equation}{Eq.}{Eqs.}

\icmltitlerunning{\AlgName: Low-Dimensional Model Embeddings for Efficient Model Exploration, Comparison, and Selection}

\begin{document}

\twocolumn[
  \icmltitle{\AlgName: Low-Dimensional Model Embeddings for \\ Efficient Model Exploration, Comparison, and Selection}
 


  \icmlsetsymbol{equal}{*}

  \begin{icmlauthorlist}
    \icmlauthor{Shivam Patel}{cmu}
    \icmlauthor{William Cocke}{cmu}
    \icmlauthor{Gauri Joshi}{cmu}
  \end{icmlauthorlist}

  \icmlaffiliation{cmu}{Carnegie Mellon University}

  \icmlcorrespondingauthor{Shivam Patel}{shivamap@andrew.cmu.edu}

  \icmlkeywords{Model Embeddings, Routing, Attention-based Encoder}

  \vskip 0.3in
]



\printAffiliationsAndNotice{}  % no special notice (required even if empty)

\begin{abstract}
The rapidly growing ecosystem of Large Language Models (LLMs) makes it increasingly challenging to manage and utilize the vast and dynamic pool of models effectively. We propose \AlgName{}, a method that produces low-dimensional vector embeddings that compactly represent a language model's capabilities across queries. \AlgName{} is an attention-based approach that generates embeddings by a deterministic forward pass over query encodings and evaluation scores via an encoder model, enabling seamless incorporation of new models to the pool and refinement of existing model embeddings without having to perform any retraining. We additionally train a correctness predictor that uses model embeddings and query encodings to achieve state-of-the-art routing accuracy on unseen queries. Experiments show that \AlgName{} needs up to $4.8\times$ fewer query evaluation samples than baselines to produce informative and robust embeddings. Moreover, the learned embedding space is geometrically meaningful: proximity reflects model similarity, enabling a range of downstream applications including model comparison and clustering, model portfolio selection, and resilient proxies of unavailable models.
\end{abstract}

\section{Introduction} \label{sec:introduction}

Large Language Models (LLMs) have been utilized for a wide range of tasks \cite{openai2024gpt4technicalreport,NEURIPS2020_1457c0d6_lms_are_few_shot_learners,bommasani2021opportunities}, including but not limited to question answering, code generation, reasoning, and domain-specific assistants \cite{hu2022lora}. Concurrently, the number of available LLMs has also grown exponentially, with hundreds of proprietary models and thousands of open-source models.\footnote{Presently, over $300\text{K}$ text-to-text models are open-sourced on \href{https://huggingface.co/models}{huggingface.co/models}} These models vary greatly in size, architecture, training data, and expertise \cite{lin2021surveytransformers_architectures}, offering opportunities to match inference queries to the best-suited model \cite{ding2024hybridllm, ong2025routellm}.
Utilizing and managing a vast pool of language models raises many practical challenges \cite{horwitz2025we_model_atlas,zhao2025surveylargelanguagemodels}. How can we compare models across sizes and architectures \cite{efficient_transformers_survey_2022}? How can we map model abilities to find similarities or differences among them 
\cite{pmlr-v97-kornblith19a_similarity_in_neuralnets}? How can we minimize redundancy in a model pool by retaining a small subset of models while preserving their aggregated abilities \cite{10.1145/3641289_llm_evaluation_survey}? How can we maintain a dynamic pool \cite{chatbot_arena} where new models are frequently added and older models are deprecated? Addressing these questions necessitates a representation that makes model capabilities easy to summarize and compare at scale.

We propose a simple yet powerful idea: represent each model by a fixed-dimensional vector that summarizes its capabilities across queries. 
We refer to this representation as a \emph{model embedding}.

\begingroup
\renewcommand\thefootnote{}\footnotetext{%
\AlgName{} implementation is publicly available at
\href{https://github.com/patel-shivam/locus_code_release}{https://github.com/patel-shivam/locus\_code\_release}.%
}
\endgroup

\paragraph{Why are model embeddings useful?}
By embedding models in a shared vector space, we can compare and organize models using simple approaches like nearest-neighbor lookup \cite{60c19788-1128-3b5f-9275-2d63cc155832_knn_orig} and clustering \cite{a_kmeans_clustering_algorithm}. This supports \emph{model discovery and organization} (e.g., finding similar models, grouping model families, and understanding joint capability profile and redundancy), \emph{portfolio selection} (choosing a smaller set of models while covering all capabilities), and \emph{monitoring} (for security purposes and tracking model drift). Model embeddings are also useful for \emph{decision making at inference time}: they can support query routing by estimating which models are likely to perform well on a given query, allowing selection of models best suited for each query.

\paragraph{Information sources for model embeddings.}
To capture capabilities of language models, we require an information source that reflects it. In the full information setting, model parameters can be directly used for generating such embeddings, but it is infeasible due to heterogeneous model sizes and architectures \cite{zhao2025surveylargelanguagemodels}  which are often proprietary \cite{grattafiori2024llama3herdmodels,geminiteam2025geminifamilyhighlycapable,openai2024gpt4technicalreport} and accessible only via APIs. An alternative is to distill model capabilities \cite{hinton2015distillingknowledgeneuralnetwork} from output logit distributions of generated tokens, which can be problematic when there are different tokenizations used across models \cite{DBLP:journals/corr/abs-2402-13116_llm_distillation}. Instead, a practical way of capturing the capabilities of diverse models is by observing their generated responses to queries, which can be assigned performance scores (e.g., correctness value) by comparison with ground truth answer \cite{hu2024routerbench} or by using LLM-as-a-judge \cite{li-etal-2025-generation_llm-as-judge}. Concretely, for a text query $x$ and model $m$, we observe a text response and compute a score $y^{(m)}(x)$, often simply the binary correctness value. For each model $m$, we aim to map the observed set of evaluations into a vector $z_m$ which summarizes its capabilities.

\paragraph{Desiderata.}
The aforementioned use cases inform key requisites of an embedding generation approach. Concretely, we seek embeddings with the following properties:
\begin{itemize}[topsep=0pt, itemsep=-2pt]
    \item \textbf{Black-box compatible:} rely only on queries and evaluation scores, without requiring access to model weights, internal activations, or logit distributions.
    \item  \textbf{Compatible with varying evaluations:} do not require all models to be evaluated on the same fixed query set to obtain comparable embeddings.
    \item \textbf{Sample efficient:} require only a few query evaluations to produce informative embeddings, which improve as more evaluations are collected.
    \item \textbf{Training-free onboarding:} adding new models should not require training model-specific parameters, nor affect existing model embeddings.
    \item \textbf{Informative of capabilities:} support accurate performance prediction on unseen queries.
    \item \textbf{Geometrically meaningful:} proximity under distance measures (e.g., cosine or Euclidean) should reflect model similarity.
\end{itemize}

\subsection{Related works.}
Prior work on finding vector representations of language models can be broadly grouped into \emph{parametric} and \emph{nonparametric} approaches. \Cref{tab:approach_qualitative_comparison} provides a conceptual comparison of existing work with our method.

\paragraph{Parametric approaches.}
A common parametric design learns \emph{per-model embeddings as trainable parameters} along with a correctness predictor trained on model--query evaluations. EmbedLLM \citep{ICLR2025_embedllm} follows this general design, while IRT-Net \citep{chen2025learningcompactrepresentationsllm_irtnet} and JE-IRT \citep{yao2025jeirtgeometriclensllm} use an item response theory formulation for interpretable predictions \citep{hambleton1991fundamentals_irt}. Although effective for correctness prediction and routing, these embeddings are learned via stochastic gradient descent (SGD) and are therefore not uniquely determined by the evaluation data: different training runs on the same evaluation data can produce varying embeddings, an issue exacerbated by overparameterization of the predictor and embeddings. This renders distance-based analyses (e.g., nearest neighbors, clustering, or similarity search) on embedding geometry brittle. Moreover, adding new models requires training new embeddings. A related parametric approach \cite{kashani-etal-2025-representing_prompt_semantic_space_embedding} utilizes matrix factorization to represent models in query encoding space. See \cref{app:regenerating_embedllm_embeddings} for additional remarks and experiments demonstrating these limitations.

\paragraph{Nonparametric approaches.}
Nonparametric methods utilize common evaluation queries to directly compute \emph{model embeddings using geometric operations}, without learning any parameters. LLM-DNA \citep{wu2025llmdnatracingmodel} concatenates query and response encodings and downprojects them using a Gaussian random matrix to obtain model embeddings. All models are required to be evaluated on the same queries, which does not generalize well to partial evaluations. Additional evaluations of queries cannot be utilized for refining the embeddings, and there is no correctness predictor that can be directly used for query routing.

\subsection{Approach Overview and Contributions}

\providecommand{\yesicon}{\textcolor{green!60!black}{\ding{51}}}   % ✓
\providecommand{\noicon}{\textcolor{red!70!black}{\ding{55}}}      % ✗
\providecommand{\maybeicon}{\textcolor{orange!80!black}{\ding{108}}} % ●

\begin{table}[t]
\centering
\scriptsize
% \tiny
\setlength{\tabcolsep}{3pt}
\renewcommand{\arraystretch}{1.1}
\caption{Qualitative comparison of \AlgName{} with prior approaches to model representations. \AlgName{} enjoys benefits of both parametric (EmbedLLM, IRT-Net) and nonparametric approaches (LLM-DNA). \yesicon{} = satisfied; \noicon{} = not satisfied}
\begin{tabular}{lccc}
\toprule
\textbf{Method} &
\begin{tabular}[c]{@{}c@{}}Training-free\\ Embeddings\end{tabular} &
\begin{tabular}[c]{@{}c@{}}Correctness\\ Prediction \end{tabular} &
\begin{tabular}[c]{@{}c@{}}Varying\\ Eval Queries\end{tabular} \\
\midrule
EmbedLLM \cite{ICLR2025_embedllm}     & \noicon    & \yesicon   & \yesicon \\
IRT-Net \cite{chen2025learningcompactrepresentationsllm_irtnet}       & \noicon    & \yesicon   & \yesicon \\
LLM-DNA \cite{wu2025llmdnatracingmodel} & \yesicon   & \noicon    & \noicon    \\
\midrule
\textbf{\AlgName~(ours)} & \yesicon & \yesicon & \yesicon \\
\bottomrule
\end{tabular}
\label{tab:approach_qualitative_comparison}
\end{table}

Motivated by strengths of both parametric and nonparametric embedding generation approaches, we propose \textbf{LO}w-dimensional embeddings for Model \textbf{C}apability \textbf{U}nder\textbf{S}tanding (\textbf{LOCUS}),
an attention-based method that processes a collection of queries evaluated on a model to generate a fixed-dimensional model embedding via an encoder $F_\theta$. After training the encoder once, generating an embedding for a model (including newly added models) is just a deterministic forward pass over its available evaluations, and the embedding can be further refined by adding more evaluations. In addition, we train a correctness predictor $G_\psi$ which processes a model embedding and a query encoding to predict the correctness of model on that query. 

\begin{figure}[t]
    \centering
    \includegraphics[width=0.7\linewidth]{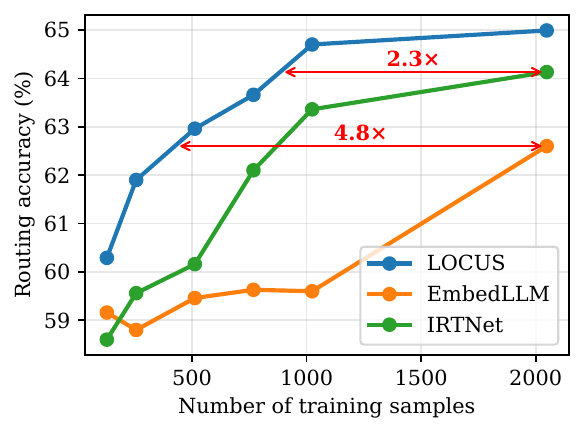}
    \caption{\textbf{Routing accuracy vs number of training samples.}  \AlgName{} uses $\mathbf{2.3-4.8}\times$ fewer training samples (number of queries on which each model is evaluated) as compared to baselines, while maintaining high routing accuracy performance.}
    \label{fig:train_sample_eff}
\end{figure}

The attention layers in \AlgName{} jointly process the available query evaluations, allowing them to extract a strong indicator of model capabilities even from limited data. We validate this insight in \cref{fig:train_sample_eff}, where \AlgName{} achieves higher routing accuracy across a wide range of training-set sizes, with up to $\mathbf{4.8\times}$ \textbf{better sample efficiency} than prior baselines. The embeddings are less prone to overfitting to a particular model’s evaluation set and remain stable as evaluation queries vary. Finally, the rich geometry of the embedding space enables several downstream tasks, including identifying model families, selecting model portfolios, and allowing resilient fallback decisions.

The remainder of the paper is organized as follows: \cref{sec:methods} introduces our approach; 
\cref{sec:experiments} reports results and analyses; and \cref{sec:conclusion} concludes our work.

\section{Using Attention for Model Capability Representation}\label{sec:methods}

In this section, we describe an attention-based encoder ($F_\theta$) for generating model embeddings from scored query evaluations, together with a query correctness predictor ($G_\psi$) that estimates a model's probability of correctness on a test query. Each query $x$ is represented by a fixed dimensional encoding/embedding $\phi(x)\in\mathbb{R}^{d_\phi}$ (e.g., generated by a pretrained sentence encoder \cite{minilm_sentence_encoder,Lan2020ALBERT}).
Let $\mathcal{Q}_m$ denote the set of queries model $m\!\in\!\mathcal{M}$ is evaluated on. Let normalized evaluation scores be $y^{(m)}(x_i)\!\in[0,1]$ (typically binary correctness $\in\!\{0,1\}$) for all queries $x_i\!\in\!\mathcal{Q}_m$. The evaluation set for model $m$ is thus
\begin{align*}
S_m \;=\; \{(\phi(x_i),\, y^{(m)}(x_i))\}_{x_i\in\mathcal{Q}_m} 
\end{align*}
with $n_m := |S_m|$ denoting the number of evaluations for model $m$.

In \cref{subsec:methods_encoder}, we describe a tokenization scheme for individual evaluations, present the standard and an efficient variant of multi-head attention over evaluation tokens, and a final aggregation layer which produces model embeddings. \Cref{subsec:methods_decoder} describes query correctness predictor, and \cref{subsec:methods_training} elaborates on the training procedure for ($F_\theta$,$G_\psi$).

\subsection{Attention-based model capability encoder $F_\theta$} \label{subsec:methods_encoder}

To process the individual query evaluations in $S_m$ for generating model embeddings, we need to convert each evaluation pair of query encoding and correctness score into a systematic representation for facilitating calculations. 
\paragraph{Continuous tokenization of evaluations.}
We convert each evaluation pair $(\phi(x_i), y^{(m)}(x_i))$ into a $d$-dimensional token $t^{(m)}_i$ using a trained MLP $h_\omega\!:\!\mathbb{R}^{d_\phi}\!\times\!\mathbb{R}\!\rightarrow\!\mathbb{R}^{1\times d}$ as:
\begin{align*} 
\fbox{ $ \;t_i^{(m)} \;=\; h_\omega\!\left(\phi(x_i),\; y^{(m)}(x_i)\right)\in\mathbb{R}^{1\times d}, \;\;$}
\end{align*}
and stack tokens as rows to form
\begin{align}\label{eq:tokenized_evaluations}
X_m^{(0)} \;=\; \begin{bmatrix} t_1^{(m)} \\ \vdots \\ t_{n_m}^{(m)}\end{bmatrix}\in\mathbb{R}^{n_m\times d}.
\end{align} 
Intuitively, $h_\omega$ acts as a continuous tokenizer mapping heterogeneous inputs (query features + score) into a common token space suitable for attention operations. We view the tokenizer $h_\omega$ as the first stage of the encoder and define $X_m^{(0)}=h_\omega(S_m)$. The subsequent attention layers then map $X_m^{(0)} \mapsto z_m$, and by slight abuse of notation we write $F_\theta(S_m)\equiv F_\theta(X_m^{(0)})$. The encoder $F_\theta$ ultimately generates a single model embedding $z_m\in\mathbb{R}^{d}$ from the query evaluations. 

We require the model embeddings to depend only on the query evaluation set $S_m$, and not on the particular order in which they are listed in $X_m^{(0)}$. To achieve this, we apply \emph{bidirectional} attention \emph{without} positional encodings. Under this design choice, shuffling the rows of $X_m^{(0)}$ only shuffles the intermediate token representations, implying permutation equivariance. We finally apply a learned-query attention aggregation step (\cref{eq:learned_query_aggregation}) which maps the last layer's output tokens into a single model embedding vector $z_m$. This operation performs aggregation without any position specific weights, resulting in a final embedding that is permutation invariant to token ordering in $X_m^{(0)}$ (\cref{eq:tokenized_evaluations}).

\begin{figure*}[htbp]
  \centering
  \begin{minipage}[t]{0.74\textwidth}
    \centering
    \includegraphics[width=\linewidth]{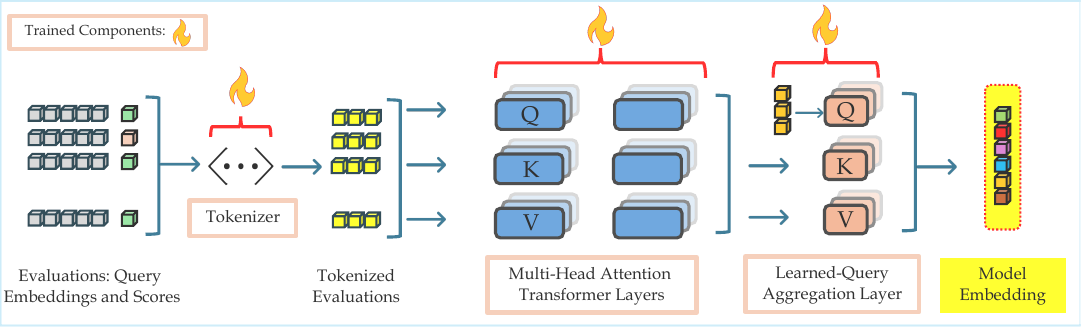}
    \subcaption{Model Embedding Generator $F_\theta$}
  \end{minipage}%
    \hspace{0.0\textwidth}
  {\color{gray}\vrule width 1pt}
  \hspace{0.0\textwidth}
  \begin{minipage}[t]{0.24\textwidth}
    \centering
    \includegraphics[width=\linewidth]{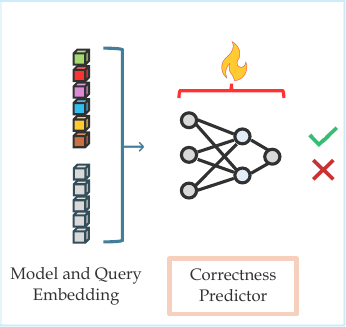}
    \subcaption{Correctness Predictor $G_\psi$}
  \end{minipage}
  \caption{\textbf{Overview of \AlgName{}.} 
  We design a model embedding generator and a correctness predictor which enables compact model representations and their capability predictions on queries. \textbf{(a) Model Embedding Generator.} Query evaluations for any language model are tokenized using the corresponding query encodings and correctness scores. Multi-Head Attention Transformer layers utilize bidirectional attention without positional encodings to capture information from the evaluations, which is finally captured by a learned query aggregation block to produce a model embedding. \textbf{(b) Correctness Predictor.} Model embedding and test query embedding are utilized by an MLP based decoder network to predict correctness probability (i.e., score) of chosen model on the query.}
  \label{fig:approach_diagram}
\end{figure*}

\paragraph{Attention over evaluation tokens.}
We denote a Transformer block \cite{NIPS2017_3f5ee243_vaswani} by $\mathrm{TBlock}(\cdot,\cdot,\cdot)$, consisting of (i) a Multi-Head Attention (\textrm{MHA}) sublayer with a residual connection and layer normalization ($\mathrm{LN}_0$), followed by (ii) a position-wise feed-forward network ($\mathrm{FFN}$, a single hidden layer \textrm{MLP} $d\!\to\! d_{\mathrm{ff}}\!\to\! d$) applied independently to each token along with another residual connection and layer normalization ($\mathrm{LN}_1$):
\begin{equation*}
\bar{Q} = \mathrm{LN}_0\Big(Q + \mathrm{MHA}(Q,K,V)\Big)\in\mathbb{R}^{n_q\times d}, 
\end{equation*}
\begin{equation*}
\mathrm{TBlock}(Q,K,V) = \mathrm{LN}_1\Big(\bar{Q} + \mathrm{FFN}(\bar{Q})\Big)\in\mathbb{R}^{n_q\times d}.
\end{equation*}

Let $X_m^{(\ell)}\in\mathbb{R}^{n_m\times d}$ denote the output tokens after $\ell$ layers of Transformer blocks ($X_m^{(0)}\in\mathbb{R}^{n_m\times d}$ are input evaluation tokens in \cref{eq:tokenized_evaluations}). Our self-attention block at layer $\ell$ over evaluation tokens is thus:
\begin{align*}
X_m^{(\ell)} \;=\; \mathrm{TBlock}\big(X_m^{(\ell-1)},\,X_m^{(\ell-1)},\,X_m^{(\ell-1)}\big)\in\mathbb{R}^{n_m\times d}.
\end{align*}

\paragraph{Latent bottleneck attention block.}
Direct self-attention over $n_m$ tokens incurs $\mathcal{O}(n_m^2)$ computational cost, which can be restrictive with large number of evaluations~$n_m$. We reduce this by introducing
$r\ll n_m$ learned latent vectors $U^{(\ell)}\in\mathbb{R}^{r\times d}$ which act as an attention bottleneck in each layer. Concretely, \emph{one latent
bottleneck attention block consists of two Transformer blocks}: a \emph{compression} block where $U^{(\ell)}$ attends to the
evaluation tokens $X_m^{(\ell-1)}$, followed by a \emph{broadcast} block in which the evaluation tokens attend to the latent outputs $H_m^{(\ell)}$:
\begin{align}
\begingroup
\setlength{\fboxsep}{3pt}
\fbox{
  \begin{minipage}{0.87\columnwidth}
  \centering
  $\begin{aligned}
  H_m^{(\ell)} \;&=\; \mathrm{TBlock}\big(U^{(\ell)},\;X_m^{(\ell-1)},\;X_m^{(\ell-1)}\big)\in\mathbb{R}^{r\times d},\\
  X_m^{(\ell)} \;&=\; \mathrm{TBlock}\big(X_m^{(\ell-1)},\;H_m^{(\ell)},\;H_m^{(\ell)}\big)\in\mathbb{R}^{n_m\times d}.
  \end{aligned}$
  \end{minipage}
}
\endgroup
\label{eq:latent_bottleneck_tblock}
\end{align}
This reduces the overall attention computation cost to $\mathcal{O}(n_m\!\times\!r)$ while still enabling full information flow through the attention layers \cite{lee2019settransformer,locatello2020objectcentriclearningslotattention_latent_bottleneck,jaegle2021perceivergeneralperceptioniterative_bottleneck_linear_attention}.

\paragraph{Learned-query attention aggregation.}
After $L$ latent-bottleneck attention blocks, we aggregate the $n_m$ output token features into a single model embedding $z_m$ using a trained query parameter $s\in\mathbb{R}^{1\times d}$, which is learned with other model parameters during training:
\begin{align}\label{eq:learned_query_aggregation}
\fbox{$\;z_m \;=\; \mathrm{TBlock}(s,\; X_m^{(L)},\; X_m^{(L)}) \in \mathbb{R}^{1\times d}.\;$}
\end{align}
This operation performs \emph{attention-based aggregation}, where a fixed, learnable query attends over a variable number of token features \cite{yang-etal-2016-hierarchical_attn_learned_aggregation,lin2017structuredselfattentivesentenceembedding_learned_aggregation,pmlr-v80-ilse18a_attention_based_multiple_input_learning,lee2019settransformer}, generating a single output vector which we denote as the model embedding. As this operation depends only on attention over the token features and does not use positional encodings, the resulting $z_m$ is invariant to permutations of the input evaluations $X^{(0)}_m$ in \cref{eq:tokenized_evaluations}.

\subsection{Query correctness predictor $G_\psi$} \label{subsec:methods_decoder}
Generated model embeddings $\{z_m\}_{m\in\mathcal{M}}$ in \cref{eq:learned_query_aggregation} are actionable for model selection only if we have a mechanism for translating the encoded information into correctness prediction of models for unseen queries. 
Such query-wise correctness prediction scores for language models can be used for a multitude of applications such as routing, confidence-based majority averaging of responses, ranking model performances etc. 
To this end, we design a correctness predictor (decoder) that processes query encodings along with model embedding vectors to return a probability of correctness value. Formally, given model embedding $z_m$ and query encoding $\phi(x)$, the decoder predicts 
\begin{align}\label{eq:correctness_probability}
\widehat{p}_\psi\left(y^{(m)}(x)\!=\!1 \Big| z_m, \phi(x)\right)=\sigma\big(G_\psi\left(z_m,\phi(x)\right)\!\big),
\end{align}
where $G_\psi$ is a lightweight MLP producing a scalar logit and $\sigma(\cdot)$ is the sigmoid function.

The overall architecture thus consists of two components: the encoder $F_\theta$ is responsible for producing a capability representation $z_m$ that is useful across queries, while the predictor $G_\psi$ combines $z_m$ with query features to output a correctness probability value.

\subsection{Training and usage}\label{subsec:methods_training}
We train the model encoder $F_\theta$ (including tokenizer $h_\omega$) and decoder $G_\psi$ (correctness predictor) jointly from a supervised dataset containing model-query evaluation pairs and observed correctness values.  
Each step samples a mini-batch of models; for each model $m$, we form (i)~an \emph{encoder input subset} $S_m^{\mathrm{enc}}$ of scored evaluations to build $z_m$, and (ii)~a \emph{decoder query batch} $S_m^{\mathrm{dec}}$ on which we apply the correctness prediction loss. Additional details provided in \cref{app:implementation_details_reproducibility_checklist}. 

\paragraph{Batch structure.}
For any model $m$, let $S_m^{\mathrm{enc}}\!\subseteq\!S_m$ be a randomly chosen subset of query evaluations, and let $z_m\!=\!F_\theta(S_m^{\mathrm{enc}})$ denote the corresponding model embedding generated.
Independently, we also randomly sample decoder queries $S_m^\mathrm{dec}\!\subseteq\!S_m$ (with target labels $y^{(m)}(x)$ for $x\!\in\!S_m^\mathrm{dec}$) and compute correctness probabilities in \cref{eq:correctness_probability} as:
\begin{align*}
\widehat{p}^{(m)}(x) \triangleq \widehat{p}_\psi\!\left(y^{(m)}(x)\!=\!1 \Big| z_m, \phi(x)\right).
\end{align*}

\paragraph{Training Loss.}
We minimize binary cross-entropy loss averaged over the models in a mini-batch and decoder  queries:
\begin{align*}
\min_{\omega,\theta,\psi}\;
\mathbb{E}_{m}\;
\mathbb{E}_{x\sim S_m}\;
\mathrm{BCE}\!\left(\widehat{p}^{(m)}(x),\, y^{(m)}(x)\right).
\end{align*}

\paragraph{Generating embeddings for new models.}
To embed a new model $m_\mathrm{new}$, we collect an evaluation set $S_{m_\mathrm{new}}^\text{enc}$, and perform a forward pass over the embedding generator to obtain $z_{m_\mathrm{new}}=F_\theta(S_{m_\mathrm{new}}^\text{enc})$. No retraining or learning any additional parameters is required at embedding time. 

\paragraph{Correctness prediction and routing.}
We estimate correctness probabilities $\widehat{p}^{(m)}(x)$ for candidate models and query $x$ by using the correctness predictor $G_\psi$. These probabilities can be used directly for ranking models, routing to a suitable model for response generation, etc.

\section{Experiments} \label{sec:experiments}
We evaluate our attention-based model embedding generator and query correctness predictor on five distinct aspects:
\begin{enumerate}[label={(\roman*)}, topsep=-2pt, itemsep=-2pt]
    \item \emph{Correctness prediction and routing accuracy}
    \item \emph{Sample Efficiency} during training and test time
    \item \emph{Robustness} of generated embeddings to changes in the evaluation set queries, both in terms of numerical performance and embedding geometry
    \item \emph{Correlation} between embedding distance and correctness similarity of models
    \item \emph{Practical utilities} enabled by model embeddings, including fallback routing and model portfolio selection.
\end{enumerate}

\paragraph{Models, tasks, and evaluation sets.}
In our experiments, we consider a set of $112$ language models \cite{ICLR2025_embedllm} spanning a wide variety of sizes and capabilities, including both general and finetuned models. The queries in our analysis are sampled from $10$ public benchmarks, and represent a real-world corpus across a variety of tasks such as math, general knowledge, reasoning etc. Additional details on language models used and query datasets are provided in \cref{app:models_and_queries_datasets}. We select the sentence encoder \texttt{all-mpnet-base-v2} \cite{song2020mpnetmaskedpermutedpretraining} as $\phi(x)$ for generating query representations in our analysis, and defer ablations on other sentence encoders to \cref{app:encoder_ablation}.

Our attention-based encoder\footnote{trained on $4096$ query evaluations for each language model, unless mentioned otherwise.} has two latent bottleneck attention blocks (\cref{eq:latent_bottleneck_tblock}) ($L\!=\!2$), $4$-headed MHA (Multi-Head Attention) sublayers, $r\!=\!64$ learned latent vectors in each block, followed by attention-based aggregation layer (\cref{eq:learned_query_aggregation}). The correctness predictor (\cref{eq:correctness_probability}) is a single hidden-layer MLP ($d_\mathrm{hidden}\!=\!64$). The model embedding dimension is set to $d\!=\!128$ (results on varying embedding dimensions deferred to \cref{app:embedding_dim_ablation}). We do not utilize any test queries for generating model embeddings; they are only used for evaluating correctness prediction and performance agreement rate of models. Further details in \Cref{app:implementation_details_reproducibility_checklist}.

\paragraph{Baselines.}
We compare against two baselines that obtain model embeddings via gradient-based training from query evaluation data: \textbf{EmbedLLM} \cite{ICLR2025_embedllm}, and \textbf{IRT-Net} \cite{chen2025learningcompactrepresentationsllm_irtnet}.

{

\begin{table}[t]
\centering
\caption{Overall routing and correctness prediction accuracy (\%) across approaches and training-set sizes (number of evaluation queries). \AlgName{} outperforms examined baselines on routing accuracy, and all approaches perform comparatively on correctness prediction accuracy. Extended results presented in \cref{app:routing_correctness_prediction}. \\\textbf{Best} and \underline{second best} emphasized in each case.
}
\label{tab:concise_routing_correctness}
\small
\setlength{\tabcolsep}{3.0pt}
\renewcommand{\arraystretch}{1.15}

\begin{tabular}{@{}l ccc ccc@{}}
\toprule

\multirow{2}{*}{\shortstack[c]{\rule{0pt}{2.7ex}Approach\\\scriptsize(\# Query Evaluations)\rule[-1.2ex]{0pt}{0pt}}} &

\multicolumn{3}{c}{Routing Accuracy (\%)} &
\multicolumn{3}{c}{Corr. Pred Acc (\%)} \\
\cmidrule(lr){2-4}\cmidrule(lr){5-7}
& 256 & 512 & 1024 & 256 & 512 & 1024 \\
\midrule

\AlgName~(Ours) 
& \textbf{61.90} & \textbf{62.97} & \textbf{64.70}
& \textbf{68.31} & \underline{68.33} & \underline{70.03} \\

\textsc{EmbedLLM}
& 58.80 & 59.47 & 59.60
& 67.33 & 68.12 & 69.47 \\

\textsc{IRT-Net}
& \underline{59.57} & \underline{60.17} & \underline{63.37}
& \underline{67.38} & \textbf{69.07} & \textbf{70.12} \\

\bottomrule
\end{tabular}
\end{table}
}

\subsection{Correctness Prediction and Routing Accuracy} \label{subsec:exp_correctness_pred_routing}

For each model $m$ and query $x$, the correctness predictor estimates
$\widehat{p}^{(m)}(x)\!=\!\sigma(G_\psi(z_m,\phi(x)))\!\in\![0,1]$. We convert the probability estimate into a binary prediction $\widehat{y}^{(m)}(x)\!=\!\mathbb{I}\{\widehat{p}^{(m)}(x)\!\geq\!0.5\}$ by thresholding, and then average the agreement indicator $\mathbb{I}\{\widehat{y}^{(m)}(x)\!=\!y^{(m)}(x)\}$ over all model-query pairs to calculate correctness prediction accuracy. Additionally, we judge whether the language model with highest predicted correctness probability $\widehat{p}^{(m)}(x)$ actually generated a correct response to query $x$, and report an average over all test queries as the overall routing accuracy.

Across training set sizes, \AlgName{} outperforms examined baselines in overall routing accuracy and correctness prediction accuracy as presented in \cref{fig:train_sample_eff,tab:concise_routing_correctness}. We attribute this to the ability of attention blocks in extracting model capability patterns through query evaluations, instead of using backpropagation to obtain model embeddings that describes observed correctness patterns. We extend further examination on numerical performance in \cref{app:routing_correctness_prediction,app:encoder_ablation,app:embedding_dim_ablation}, and wall clock time complexity in \cref{app:time_complexity_locus}.

\subsection{Sample Efficiency} \label{subsec:exp_sample_efficiency}
A key motivation for model embeddings is \emph{sample-efficient learning and new model onboarding}: we aim to train a general-purpose model capability encoder from limited evaluations, and efficiently embed previously unseen models using small number of query evaluations.

\begin{figure}[t]
    \centering
    \includegraphics[width=0.8\linewidth]{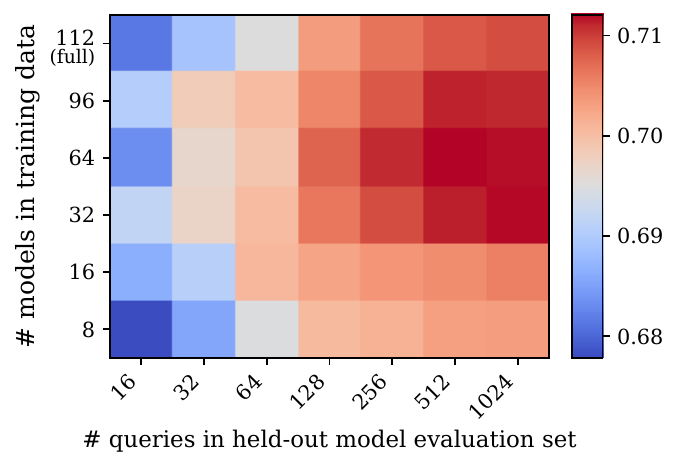}
    \caption{\textbf{New LLM onboarding.} Correctness prediction accuracy on $16$ held-out models: with varying number of query evaluations for embedding held-out models and varying number of models present in training data (top row indicates $(F_\theta, G_\psi)$ trained on the full set of all language models). Only a few queries ($\approx$$128$) are required to generate informative embeddings for unseen models, and there is a negligible ($< 1\%$) accuracy gap between using an encoder trained on the full (top row) versus a partial set of models.  
    }

    \label{fig:sample_eff_ood_test}
\end{figure}

\paragraph{Training-time sample efficiency.}
\Cref{fig:train_sample_eff} reports routing accuracy as a function of the number of query evaluations per model used to train
$(F_\theta,G_\psi)$ (i.e., $n_m=|S_m|$). Our method reaches strong correctness prediction and routing performance with substantially fewer training evaluations, resulting in \textbf{upto} $\mathbf{4.8 \times}$ \textbf{sample efficiency} as compared to existing baselines. Notably, our training procedure does not require all models to share the same evaluation queries, nor to be evaluated on the same number of queries, further increasing practicality of our approach.

\paragraph{Test-time embedding efficiency (new model onboarding).}
To measure sample efficiency for generating embeddings for unseen models, we train $(F_\theta,G_\psi)$ on query evaluation data from \# models $\in\{8,16,\ldots,92\}$ ($1024$ evaluations each) and then embed $16$ held-out models using \# query evaluations $\in\{16,32,\ldots,1024\}$ (details in \cref{app:models_and_queries_datasets}). \Cref{fig:sample_eff_ood_test} shows that as few as $\approx$$128$ evaluations are sufficient to obtain informative embeddings. We also observe negligible gap ($<1\%$) between correctness prediction accuracy using partial data trained versus full data trained $(F_\theta,G_\psi)$, highlighting generalization to unseen models.

\begin{figure}[t]
    \centering
    \includegraphics[width=\linewidth]{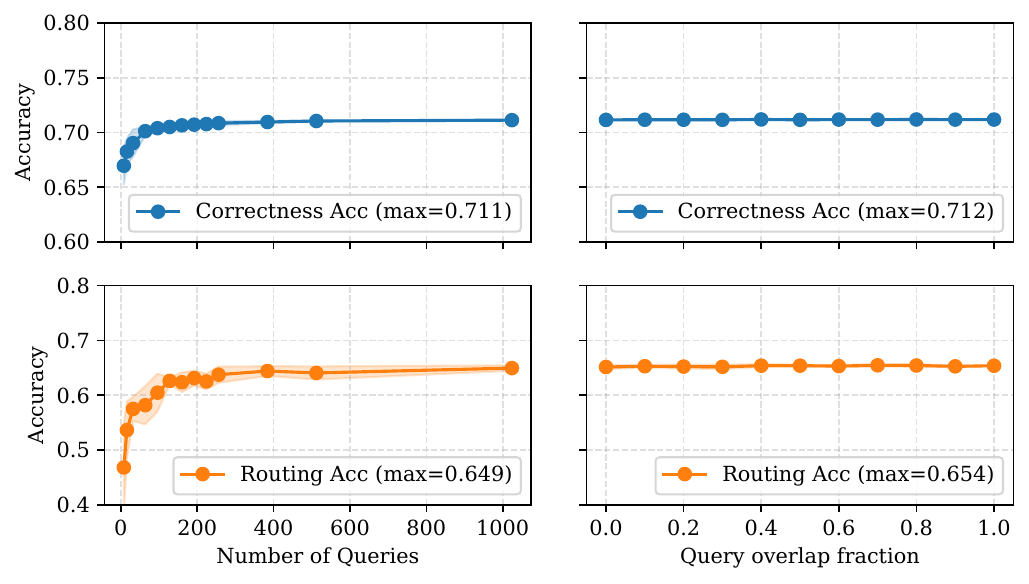}
    \caption{\textbf{Robustness to Varying Query Evaluations.} Accuracy for correctness prediction (top) and routing (bottom) as a function of (left) the number of evaluations used to construct embeddings and (right) the overlap fraction with a reference evaluation set. 
    Performance saturates with as few as $128\text{-}256$ evaluations, and remains stable across overlap fractions and query subsampling, indicating robustness to evaluation queries.}
\label{fig:perf_vs_size_and_overlap}
\end{figure}

\subsection{Robustness to Varying Query Evaluations} \label{subsec:exp_embedding_stability_robustness}

We test how sensitive model embeddings are to the query evaluation set used as input to the encoder. 
Using a pretrained encoder
$F_\theta$ trained from evaluations on the full model pool, we first compute \emph{reference embeddings}
$z_m^{\mathrm{ref}}$ for each model $m$ from a fixed reference query evaluation set of size $n_{\mathrm{ref}}$ (e.g., $4096$).
We then perturb the input evaluations in two ways: changing \textbf{which} queries are used (fixed number but different queries) and
changing \textbf{how many} queries are used (subsampling from reference evaluation set), and then analyze the stability of embeddings via routing/correctness prediction and embedding geometry.

\begin{figure}[t]
    \centering
    \includegraphics[width=0.8\linewidth]{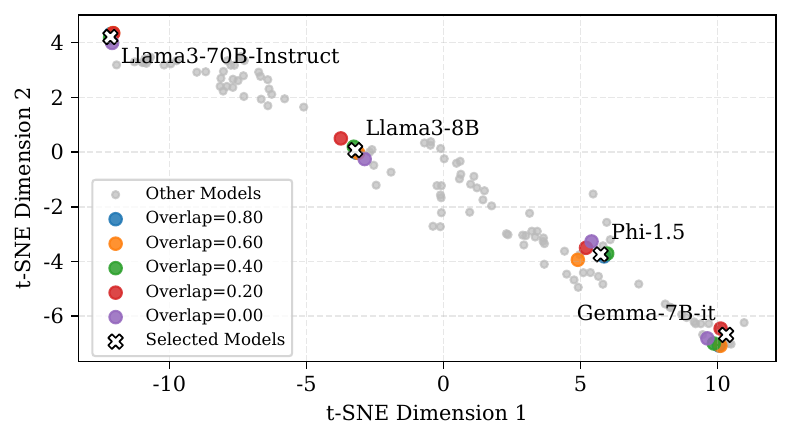}

    \includegraphics[width=0.8\linewidth]{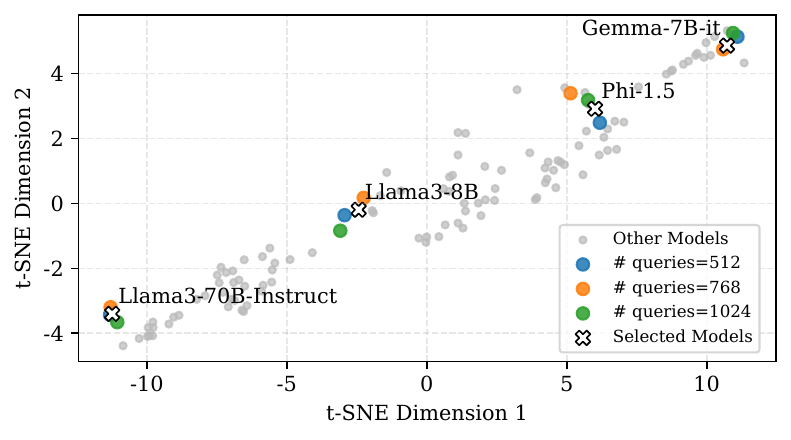}

    \caption{\textbf{Robustness to varying query evaluations.}
    Embeddings for selected models recomputed from evaluation sets with {(top)} varying overlap relative to the reference set and
    {(bottom)} evaluation queries subsampled from the reference set, both visualized with t-SNE.
    Gray points denote other models in the pool.}
    \label{fig:embedding_robustness_tsne}
\end{figure}

\paragraph{Robustness to the choice of evaluation queries.}
Keeping the number of queries in the evaluation set fixed at $n_{\mathrm{ref}}$, we resample alternate evaluation sets whose overlap with the reference
set is $\alpha\in[0,1]$. We then recompute embeddings for all models, and calculate numerical metrics using the new embeddings. \Cref{fig:perf_vs_size_and_overlap} shows that correctness prediction and routing accuracy remain stable across overlap fractions, and \cref{fig:embedding_robustness_tsne} illustrates that
the regenerated embeddings stay close to originally generated embeddings $z_m^{\mathrm{ref}}$ (downprojection via t-SNE \cite{JMLR:v9:vandermaaten08a_tsne}).

\paragraph{Robustness to the number of evaluation queries.}
We also subsample $n$ evaluations from the reference set of query evaluations (e.g., $n\in\{64,128,256,\ldots,n_{\mathrm{ref}}\}$) and recompute
embeddings. Numerical performance saturates quickly: embeddings formed from only $\approx128$--$256$ evaluations typically match
the routing and correctness prediction accuracy obtained with larger sets (\cref{fig:perf_vs_size_and_overlap}). The corresponding
t-SNE projections of embeddings in \cref{fig:embedding_robustness_tsne} further suggest geometric stability under subsampling evaluation queries.

\paragraph{Implications for model fingerprinting (security).}
Robustness to query evaluations provides a preliminary fingerprinting signal: if a new model’s embedding repeatedly maps near the same reference model across independently sampled evaluation sets, it suggests potential model duplication and can be flagged for audit (see \cref{subapp:embedding_convergence_security} for additional experiments) \cite{zeng2024huref_fingerprinting,tsai2025roflrobustfingerprintinglanguage}.

\subsection{Embedding Geometry reflects Model Similarity} 
\label{subsec:exp_behavioural_similarity}
\begin{figure}[t]
    \includegraphics[width=\linewidth]{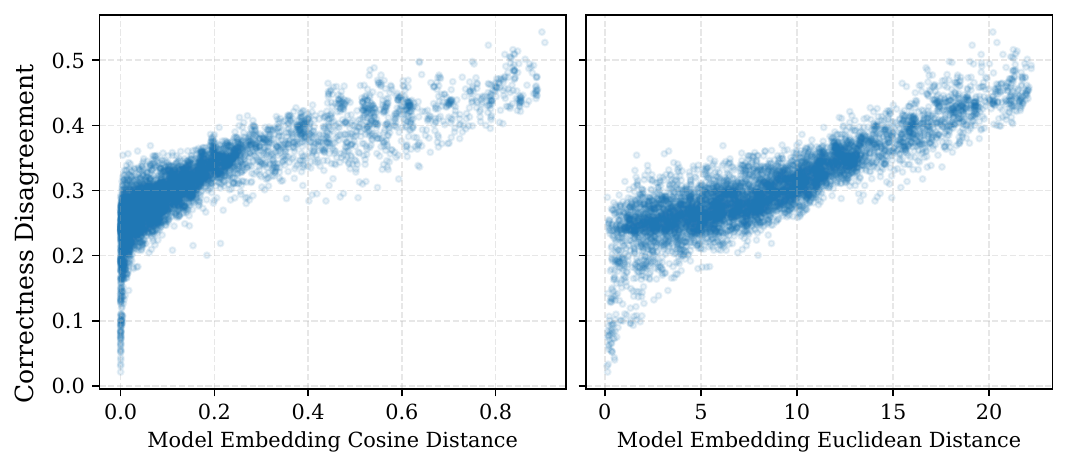}
  \caption{\textbf{Embedding distance versus correctness disagreement for pairs of models.} Distances in embedding space exhibit strong positive correlation with disagreement in observed correctness outcomes (e.g., Pearson's correlation is $0.85$ using cosine distance and $0.89$ using Euclidean distance between embeddings), indicating that distances among embeddings express model similarity.}
  \label{fig:performance_distance_correlation}
\end{figure}

We next test whether distances in the learned embedding space track differences between model capabilities. Concretely, for every model
pair $(m_1,m_2)\in \mathcal{M}\!\times\!\mathcal{M}$ we compute (i) an embedding-space distance $d_z(z_{m_1},z_{m_2})$ (cosine and Euclidean), and (ii) their correctness
\emph{disagreement rate} on a common test query set, defined as the fraction of queries on which their binary correctness labels differ
(i.e., the normalized Hamming distance, or $1-$agreement). \Cref{fig:performance_distance_correlation,tab:performance_distance_correlation}
shows a strong positive relationship between embedding distance and disagreement across all correlation measures: for example, Pearson
correlation is $0.845$ (cosine) and $0.887$ (Euclidean), with similarly high rank correlations (Spearman $\approx0.88$ and Kendall
$\approx0.71$). These results indicate that embedding distances convey meaningful information about model similarity.

\begin{table}[t]
\centering
\caption{\textbf{Correlation between embedding distance and correctness disagreement} (normalized Hamming distance between correctness labels) across all model pairs.}
\label{tab:performance_distance_correlation}
\small
\setlength{\tabcolsep}{4.0pt}
\begin{tabular}{@{}lcc@{}}
\toprule
Correlation & Cosine distance & Euclidean distance \\
\midrule
Pearson $\rho$   & $0.845$ & $0.887$ \\
Spearman $r_s$   & $0.886$ & $0.876$ \\
Kendall $\tau$   & $0.714$ & $0.702$ \\
\bottomrule
\end{tabular}
\end{table}

\begin{figure}[t]
    \centering
    \begin{subfigure}[t]{0.34\linewidth}
        \centering
        \includegraphics[width=\linewidth]{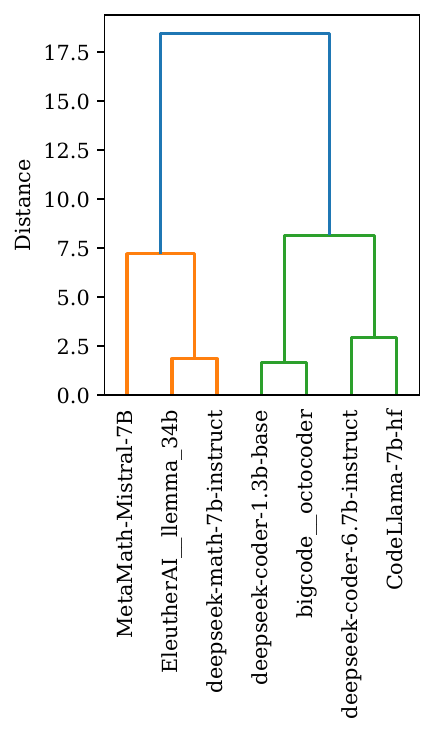}
    \end{subfigure}\hfill
    \begin{subfigure}[t]{0.64\linewidth}
        \centering
        \includegraphics[width=\linewidth]{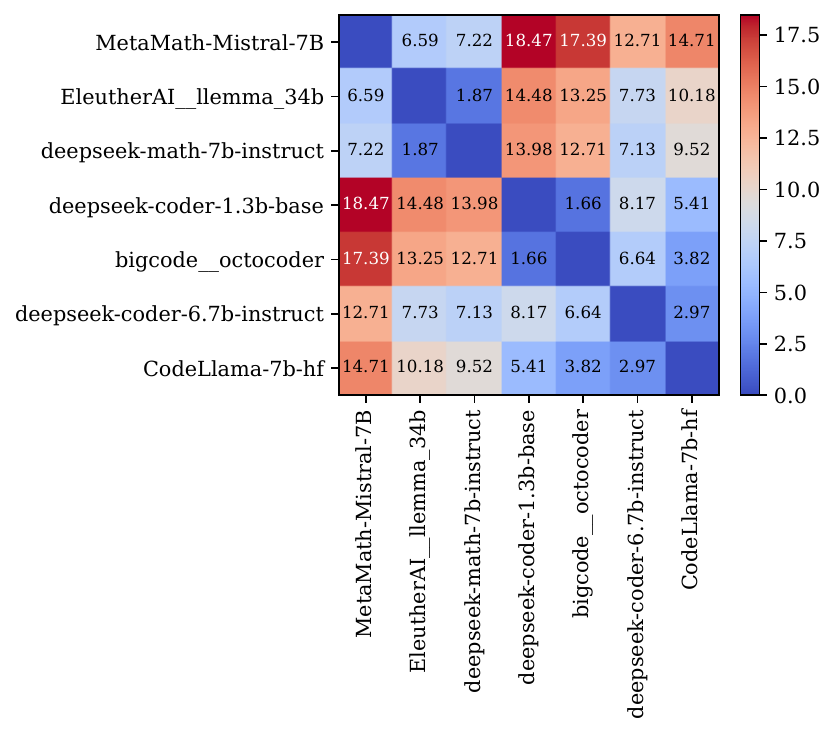}
    \end{subfigure}
    \caption{\textbf{Identifying model families.} Hierarchical clustering on model embeddings meaningfully separates model families, illustrated here for math (orange) and code (green) families. 
    }
    \label{fig:models_family_heirarchical_clustering}
\end{figure}

\paragraph{Recovering model families via hierarchical clustering.}
We perform hierarchical clustering \cite{10.1145/3321386_hierarchical_clustering} of models using pairwise distances (Euclidean distance) between embeddings.
The dendrogram and distance heatmap in \cref{fig:models_family_heirarchical_clustering} reveal coherent groupings that align with known model
specializations (e.g., math- and code-finetuned families), suggesting that the embedding geometry also captures shared capability structure at the level of model families rather than just individual models.

\subsection{Practical utilities enabled by model embeddings}\label{subsec:exp_practical_utility}
We finally demonstrate that compact, stable capability embeddings enable useful operations beyond standard routing or correctness prediction tasks.

\paragraph{Nearest neighbors as proxies under model unavailability.}

\begin{figure}[t]
    \centering
    \includegraphics[width=0.9\linewidth]{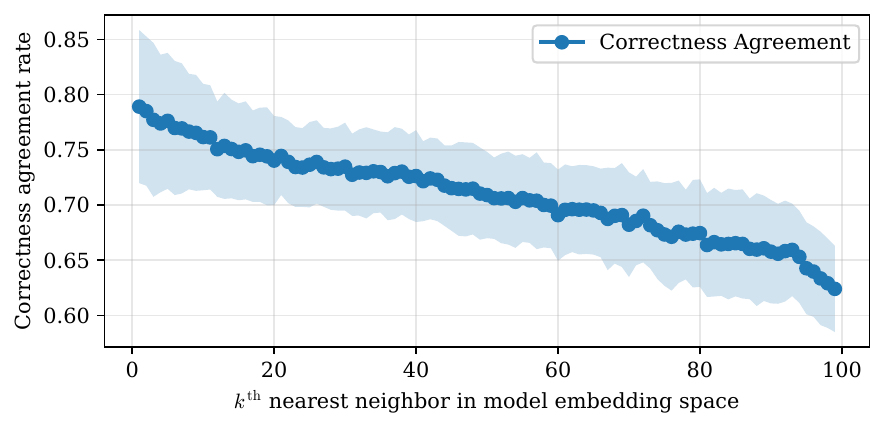}
    \caption{\textbf{Nearest-neighbor model proxies.} Average correctness agreement between a model and its $k$-th nearest
    neighbor in embedding space (averaged over models). Closest neighbors exhibit the highest agreement, which decays gradually for further neighbors. 
    }
    \label{fig:nearest_neighbor_proxies}
\end{figure}

In practice, a preferred model for generating response to a query can be inaccessible due to capacity limits, hardware failure, policy constraints etc. In such cases, it is useful to identify \emph{proxy} models with close matching capabilities. We therefore test whether nearest neighbors in embedding space can serve as reliable proxies. For each model $m$, we rank its neighbors by embedding distance (e.g., cosine distance) and calculate correctness agreement on a common set of queries. \Cref{fig:nearest_neighbor_proxies} reports the average agreement of $k^\text{th}$ neighbor across all models in the pool, with closest neighbor \textbf{correctness agreement as high as} $\mathbf{79\%}$ and decreasing gradually with further neighbors (larger $k$). 
We extend the implications of this to resilient routing in  \Cref{subapp:resilient_routing}. 

\paragraph{Model portfolio selection from embeddings.}
We evaluate whether the \emph{model embedding space} can support practical deployment planning: selecting a \emph{portfolio} of models such that routing among the selected models retains high accuracy on queries as compared to the full model pool (i.e., selecting a subset of models to eliminate redundancy in capabilities across models while maintaining performance). 

\begin{figure}[t]
    \centering
    \includegraphics[width=\linewidth]{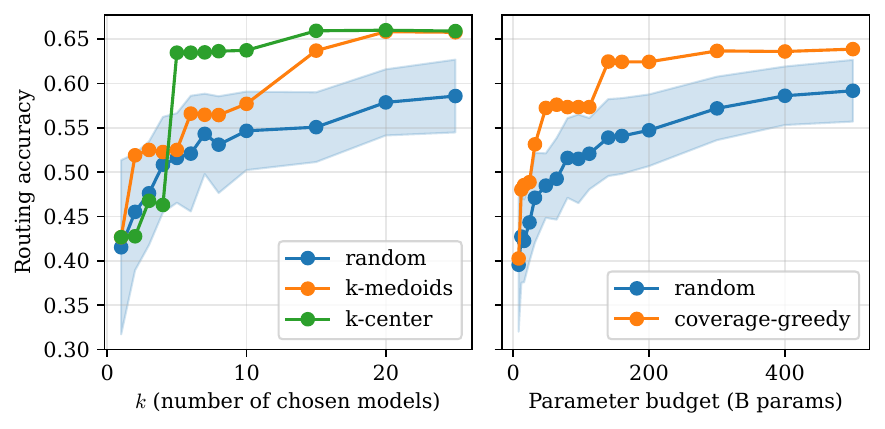}
    \caption{\textbf{Model Portfolio Selection.} Using embeddings to choose a model subset that maximizes routing accuracy. \emph{Left (count-constrained):} \texttt{k-center} and \texttt{k-medoids} maximize embedding-space coverage.
    \emph{Right (parameter-budget):} under a total-parameter budget, \texttt{coverage-greedy} selects embeddings to maximize coverage while prioritizing smaller models. 
    }
    \label{fig:model_portfolio_joint}
\end{figure}

\emph{Count-constrained portfolios.}
When limiting deployment to $k$ models, we select diverse portfolios by maximizing coverage in embedding space using \texttt{k-center} (minimizes the maximum distance between all models' embeddings to those in the chosen portfolio) \cite{GONZALEZ1985293_k_centers} and \texttt{k-medoids} (minimize average distance of all models' embeddings to models in chosen portfolio) \cite{KaufmanRousseeuw1990PAM_k_medoids}. 
We observe that a \textbf{$\mathbf{15}$-model portfolio captures full routing accuracy} of $112$ models in the pool.
Detailed algorithm is deferred to \cref{subsubapp:count_constrained_portfolio_selection}. We report routing accuracy restricted to models in the portfolio as $k$ varies (\cref{fig:model_portfolio_joint}).

\emph{Parameter-budget portfolios.}
We solve a total-parameter-budget-constrained optimization problem (\texttt{coverage-greedy}): select a model subset which covers the model embedding space while respecting the overall parameter budget (\cref{subsubapp:parameter_constrained_portfolio_selection}). \Cref{fig:model_portfolio_joint} depicts strong routing performance with relatively small parameter budget (full pool of $112$ models has a total of $1930$B parameters).

\paragraph{Additional utilities.}
Beyond model capability proxies and portfolio selection, model embeddings enable other practical operations such as resilient routing, retrieving a real model that best matches a desired capability profile etc. We provide further discussion on additional use cases in \cref{app:additional_utility_embeddings}.

\section{Concluding Remarks }
\label{sec:conclusion}
In this paper, we designed \AlgName{}, a novel attention-based framework that projects Large Language Models into a low-dimensional embedding space based on their inference capabilities. Unlike prior approaches,
our method supports seamless, training-free onboarding of new models without disturbing existing embeddings, while demonstrating superior sample efficiency, requiring up to $4.8 \times$ fewer query evaluations to generate robust representations. Beyond standard performance prediction, we demonstrated that the learned embedding geometry effectively captures behavioral similarity, enabling diverse downstream utilities such as optimal model portfolio selection and resilient fallback routing under model unavailability. As the ecosystem of proprietary and open-source models continues to expand, such capability-oriented embeddings offer a scalable, geometrically meaningful foundation for efficient model management and discovery. We leave extensions to multimodal models and adaptive query evaluations for generating embeddings as future directions.

\newpage

\section*{Acknowledgments}
This work was partially supported by NSF grants CCF 2045694, CNS-2112471, CPS-2111751, ONR grant N00014-23-1-2149, and an AI2C Seed grant. This work used Bridges-2 GPU at the Pittsburgh Supercomputing Center through allocation CIS250429 from the Advanced Cyberinfrastructure Coordination Ecosystem: Services \& Support (ACCESS) program, which is supported by NSF grants \#2138259, \#2138286, \#2138307, \#2137603, and \#2138296 \citep{access}. 

\section*{Impact Statement}
This paper presents work whose goal is to advance the field of Machine Learning. In particular, we design low-dimensional model embeddings that can be used to efficiently manage and utilize large model pools. There are many potential societal consequences of our work, none of which we feel must be specifically highlighted here.

\newpage
\bibliography{references}
\bibliographystyle{icml2026}

%%%%%%%%%%%%%%%%%%%%%%%%%%%%%%%%%%%%%%%%%%%%%%%%%%%%%%%%%%%%%%%%%%%%%%%%%%%%%%%
%%%%%%%%%%%%%%%%%%%%%%%%%%%%%%%%%%%%%%%%%%%%%%%%%%%%%%%%%%%%%%%%%%%%%%%%%%%%%%%
% APPENDIX
%%%%%%%%%%%%%%%%%%%%%%%%%%%%%%%%%%%%%%%%%%%%%%%%%%%%%%%%%%%%%%%%%%%%%%%%%%%%%%%
%%%%%%%%%%%%%%%%%%%%%%%%%%%%%%%%%%%%%%%%%%%%%%%%%%%%%%%%%%%%%%%%%%%%%%%%%%%%%%%
\newpage
\appendix
\onecolumn
\appendix
\onecolumn
\crefalias{section}{appendix}
\crefalias{subsection}{appendix}
\crefalias{subsubsection}{appendix}

\section{Language Models and Query Datasets}\label{app:models_and_queries_datasets}
\begin{figure}[h]
    \centering
    \includegraphics[width=0.8\linewidth]{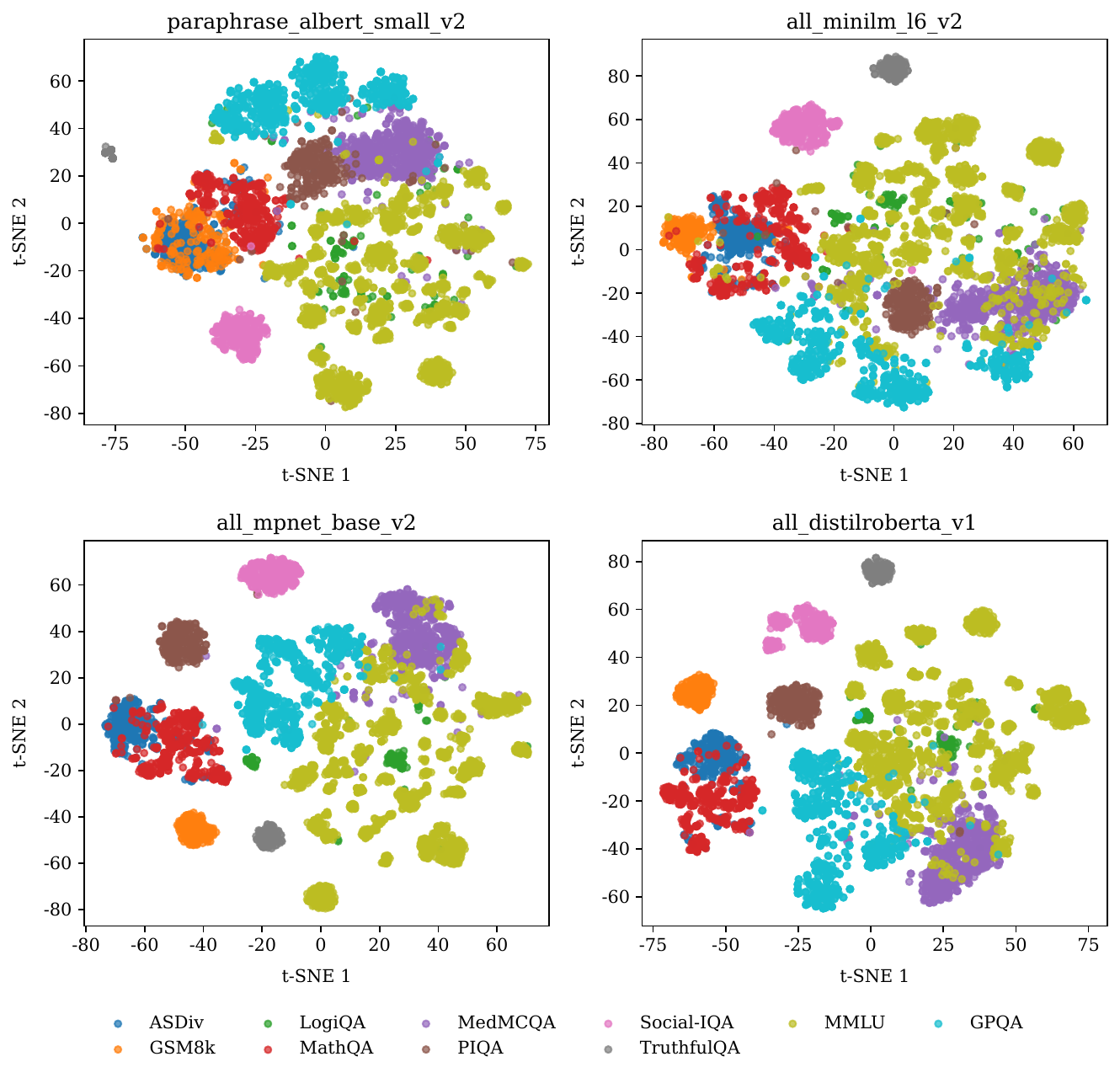}
    \caption{Query encodings visualized using t-SNE \cite{JMLR:v9:vandermaaten08a_tsne}, colored by dataset. Queries from the same dataset represent similar tasks, and their sentence encodings are placed close by in the encoding space. Plots presented for four different sentence encoders \texttt{all\_mpnet\_base\_v2} \cite{song2020mpnetmaskedpermutedpretraining}, \texttt{paraphrase\_albert\_small\_v2} \cite{Lan2020ALBERT}, \texttt{all\_minilm\_l6\_v2} \cite{minilm_sentence_encoder}, and \texttt{all\_distilroberta\_v1} \cite{sanh2020distilbertdistilledversionbert}.}
    \label{fig:query_encodings_tsne}
\end{figure}

We use the model-query evaluations presented in \citet{ICLR2025_embedllm}. 
The queries are sampled from ten public benchmarks, representing a wide variety of real-world prompts and tasks: \textbf{MathQA} \cite{amini-etal-2019-mathqa}, \textbf{LogiQA} \cite{ijcai2020p501_logiqa}, \textbf{MedMCQA} \cite{pmlr-v174-pal22a_medmcqa}, \textbf{PIQA} \cite{Bisk_Zellers_Lebras_Gao_Choi_2020_piqa}, \textbf{TruthfulQA} \cite{lin-etal-2022-truthfulqa}, \textbf{MMLU} \cite{hendrycks2021measuring_mmlu}, \textbf{GSM8K}\cite{cobbe2021trainingverifierssolvemath_gsm8k}, \textbf{GPQA}\cite{rein2024gpqa_dataset}, \textbf{ASDiv}\cite{miao-etal-2020-diverse_asdiv}, \textbf{SocialQA} \cite{sap-etal-2019-social}. Semantic query encodings generated by sentence encoders are presented in \cref{fig:query_encodings_tsne} for four choices of sentence encoders, colored by dataset. 

The 112 language models used are presented in \cref{tab:language_model_names} with their HuggingFace identifiers \cite{wolf2020huggingfacestransformersstateoftheartnatural}. For new model onboarding analysis as presented in \cref{subsec:exp_sample_efficiency,fig:sample_eff_ood_test}, we consider the following sixteen models as held-out from training (seen only during test time): {\texttt{FelixChao/llama2-13b-math1.2},
\texttt{MaziyarPanahi/WizardLM-Math-70B-v0.1},
\texttt{01-ai/Yi-6B-200K},
\texttt{WizardLM/WizardLM-70B-V1.0},
\texttt{bigscience/bloom-7b1},
\texttt{sail/Sailor-7B},
\texttt{codellama/CodeLlama-13b-Instruct-hf},
\texttt{Writer/palmyra-med-20b},
\texttt{Qwen/Qwen1.5-0.5B-Chat},
\texttt{databricks/dolly-v2-12b},
\texttt{nomic-ai/gpt4all-13b-snoozy},
\texttt{stabilityai/stablelm-tuned-alpha-7b},
\texttt{AdaptLLM/medicine-chat},
\texttt{AdaptLLM/medicine-LLM},
\texttt{EleutherAI/pythia-12b},
\texttt{Q-bert/Optimus-7B}}.

\newpage

\begin{table}[ht]
\centering
\caption{Language models utilized in our experiments, as presented in \citet{ICLR2025_embedllm}. All 112 language models are denoted by their HuggingFace identifiers \cite{wolf2020huggingfacestransformersstateoftheartnatural}.}
\label{tab:language_model_names}
\scriptsize
\setlength{\tabcolsep}{6pt}
\renewcommand{\arraystretch}{1.05}
\begin{tabular}{cc}
\hline
\texttt{Qwen/Qwen1.5-7B-Chat} & \texttt{ConvexAI/Luminex-34B-v0.1} \\
\texttt{lmsys/vicuna-13b-v1.5} & \texttt{deepseek-ai/deepseek-math-7b-instruct} \\
\texttt{TigerResearch/tigerbot-13b-base} & \texttt{ConvexAI/Luminex-34B-v0.2} \\
\texttt{berkeley-nest/Starling-LM-7B-alpha} & \texttt{EleutherAI/llemma\_7b} \\
\texttt{CultriX/NeuralTrix-bf16} & \texttt{SciPhi/SciPhi-Mistral-7B-32k} \\
\texttt{TheBloke/tulu-30B-fp16} & \texttt{lmsys/vicuna-33b-v1.3} \\
\texttt{scb10x/typhoon-7b} & \texttt{mlabonne/AlphaMonarch-7B} \\
\texttt{mistralai/Mistral-7B-Instruct-v0.1} & \texttt{01-ai/Yi-34B-Chat} \\
\texttt{meta-llama/Llama-2-13b-chat-hf} & \texttt{eren23/ogno-monarch-jaskier-merge-7b-OH-PREF-DPO} \\
\texttt{ibivibiv/alpaca-dragon-72b-v1} & \texttt{golaxy/gowizardlm} \\
\texttt{codellama/CodeLlama-34b-Instruct-hf} & \texttt{OpenBuddy/openbuddy-codellama2-34b-v11.1-bf16} \\
\texttt{deepseek-ai/deepseek-coder-1.3b-base} & \texttt{Neko-Institute-of-Science/pygmalion-7b} \\
\texttt{cognitivecomputations/yayi2-30b-llama} & \texttt{meta-llama/LlamaGuard-7b} \\
\texttt{NousResearch/Nous-Hermes-13b} & \texttt{tiiuae/falcon-40b-instruct} \\
\texttt{meta-llama/Llama-2-7b-chat-hf} & \texttt{mosaicml/mpt-7b-chat} \\
\texttt{Qwen/Qwen1.5-32B-Chat} & \texttt{NousResearch/Nous-Hermes-2-Yi-34B} \\
\texttt{deepseek-ai/deepseek-coder-6.7b-instruct} & \texttt{google/gemma-7b-it} \\
\texttt{EleutherAI/llemma\_34b} & \texttt{zhengr/MixTAO-7Bx2-MoE-v8.1} \\
\texttt{yam-peleg/Experiment26-7B} & \texttt{meta-llama/Meta-Llama-3-8B} \\
\texttt{mosaicml/mpt-30b-instruct} & \texttt{fblgit/UNA-SimpleSmaug-34b-v1beta} \\
\texttt{FelixChao/vicuna-7B-physics} & \texttt{TheBloke/koala-13B-HF} \\
\texttt{meta-llama/Meta-Llama-3-70B} & \texttt{Plaban81/Moe-4x7b-math-reason-code} \\
\texttt{meta-math/MetaMath-Mistral-7B} & \texttt{BioMistral/BioMistral-7B} \\
\texttt{FelixChao/Scorpio-7B} & \texttt{SciPhi/SciPhi-Self-RAG-Mistral-7B-32k} \\
\texttt{microsoft/phi-2} & \texttt{CausalLM/34b-beta} \\
\texttt{meta-llama/Meta-Llama-3-70B-Instruct} & \texttt{meta-math/MetaMath-Llemma-7B} \\
\texttt{lmsys/vicuna-7b-v1.5-16k} & \texttt{cloudyu/Mixtral\_11Bx2\_MoE\_19B} \\
\texttt{Qwen/Qwen1.5-4B-Chat} & \texttt{FelixChao/vicuna-7B-chemical} \\
\texttt{HuggingFaceH4/zephyr-7b-beta} & \texttt{OpenAssistant/oasst-sft-4-pythia-12b-epoch-3.5} \\
\texttt{BioMistral/BioMistral-7B-DARE} & \texttt{Biomimicry-AI/ANIMA-Nectar-v2} \\
\texttt{microsoft/phi-1\_5} & \texttt{meta-llama/Meta-Llama-Guard-2-8B} \\
\texttt{rishiraj/CatPPT-base} & \texttt{kyujinpy/Sakura-SOLRCA-Math-Instruct-DPO-v1} \\
\texttt{meta-llama/Meta-Llama-3-8B-Instruct} & \texttt{google/gemma-2b-it} \\
\texttt{upstage/SOLAR-10.7B-Instruct-v1.0} & \texttt{CorticalStack/pastiche-crown-clown-7b-dare-dpo} \\
\texttt{01-ai/Yi-6B} & \texttt{codefuse-ai/CodeFuse-DeepSeek-33B} \\
\texttt{abhishek/zephyr-beta-math} & \texttt{bardsai/jaskier-7b-dpo-v5.6} \\
\texttt{allenai/tulu-2-dpo-70b} & \texttt{Harshvir/Llama-2-7B-physics} \\
\texttt{lmsys/vicuna-13b-v1.5-16k} & \texttt{shleeeee/mistral-ko-tech-science-v1} \\
\texttt{JaeyeonKang/CCK\_Asura\_v1} & \texttt{codellama/CodeLlama-7b-hf} \\
\texttt{Nexusflow/Starling-LM-7B-beta} & \texttt{microsoft/Orca-2-13b} \\
\texttt{Neko-Institute-of-Science/metharme-7b} & \texttt{bigcode/octocoder} \\
\texttt{PharMolix/BioMedGPT-LM-7B} & \texttt{SUSTech/SUS-Chat-34B} \\
\texttt{kevin009/llamaRAGdrama} & \texttt{meta-llama/Llama-2-70b-chat-hf} \\
\texttt{TheBloke/CodeLlama-70B-Instruct-AWQ} & \texttt{openchat/openchat\_3.5} \\
\texttt{dfurman/HermesBagel-34B-v0.1} & \texttt{project-baize/baize-v2-13b} \\
\texttt{augmxnt/shisa-base-7b-v1} & \texttt{lmsys/vicuna-7b-v1.5} \\
\texttt{Intel/neural-chat-7b-v3-3} & \texttt{AdaptLLM/medicine-LLM-13B} \\
\texttt{openchat/openchat-3.5-0106} & \texttt{deepseek-ai/deepseek-llm-67b-chat} \\
\texttt{FelixChao/llama2-13b-math1.2} & \texttt{MaziyarPanahi/WizardLM-Math-70B-v0.1} \\
\texttt{01-ai/Yi-6B-200K} & \texttt{WizardLM/WizardLM-70B-V1.0} \\
\texttt{bigscience/bloom-7b1} & \texttt{sail/Sailor-7B} \\
\texttt{codellama/CodeLlama-13b-Instruct-hf} & \texttt{Writer/palmyra-med-20b} \\
\texttt{Qwen/Qwen1.5-0.5B-Chat} & \texttt{databricks/dolly-v2-12b} \\
\texttt{nomic-ai/gpt4all-13b-snoozy} & \texttt{stabilityai/stablelm-tuned-alpha-7b} \\
\texttt{AdaptLLM/medicine-chat} & \texttt{AdaptLLM/medicine-LLM} \\
\texttt{EleutherAI/pythia-12b} & \texttt{Q-bert/Optimus-7B} \\
\hline
\end{tabular}
\end{table}

\newpage
\section{Extended Numerical Comparison with Baselines} \label{app:routing_correctness_prediction}

We present the extended numerical results on routing accuracy (\cref{tab:routing_acc_by_dataset_full}) and correctness prediction (\cref{tab:correctness_pred_by_dataset_full}) of \AlgName{}, compared against three other baselines: EmbedLLM \cite{ICLR2025_embedllm}, EmbedLLM+ (more hidden layers in decoder with same architecture as original \cite{ICLR2025_embedllm}), and IRT-Net \cite{chen2025learningcompactrepresentationsllm_irtnet}. We additionally  present correctness prediction and routing accuracy of $K$-Means and $k$NN based approaches \cite{hu2024routerbench,jitkrittum2025universalmodelroutingefficient, patel2025proxrouterproximityweightedllmquery,zhang2025avengerssimplerecipeuniting}. $K$-Means routers perform clustering over query encodings and assign average model correctness on closest cluster for test queries as the predicted correctness probability, whereas $k$NN routers average model correctness over nearest $k$ queries in encoding space as the prediction for a test query. 

We observe that across all model embedding based approaches, \AlgName{} exhibits the highest overall routing accuracy (\cref{tab:routing_acc_by_dataset_full}) across all choices of number of training samples (number of queries each model is evaluated on). This is succinctly described in \cref{fig:train_sample_eff} as well. For correctness prediction accuracy (\cref{tab:correctness_pred_by_dataset_full}) we find that all approaches perform similarly in terms of overall correctness prediction accuracy. Among nonparametric routing techniques, $K$-Means outperforms other approaches on certain choices of training data size, but we highlight that we cannot generate model embeddings and support their utilities via $K$-Means clustering based routers.

\begin{table*}[htbp]
\centering
\caption{Overall and per-dataset Routing Accuracy, across approaches and training-set sizes. We compare against model embedding generation approaches EmbedLLM \cite{ICLR2025_embedllm}, EmbedLLM+ (with more hidden layers than original \cite{ICLR2025_embedllm}) and IRT-Net \cite{chen2025learningcompactrepresentationsllm_irtnet}. We also present performance of embedding free routers ($K$-Means and $k$NN are popular nonparametric routers) as strong baselines on routing tasks. Overall across all queries, \AlgName{} outperforms all model embedding based approaches, and also beats $k$NN router whereas $K$-Means acts as a strong routing baseline on some choices of training data size.}
\label{tab:routing_acc_by_dataset_full}
\scriptsize
\setlength{\tabcolsep}{2.5pt}      
\renewcommand{\arraystretch}{1.15}

\begin{tabular}{@{}l c c *{10}{c}@{}}
\toprule
\multirow{2}{*}{\textbf{Approach}} &
\multirow{2}{*}{\textbf{Train samples}} &
\multicolumn{11}{c}{\textbf{Routing Accuracy (\%)}} \\
\cmidrule(lr){3-13}
& & \textbf{Overall} & MathQA & LogiQA & MedQA & PIQA & TruthQA & MMLU & GSM8k & GPQA & ASDiv & SoQA \\
\midrule

\multirow{3}{*}{\textbf{\AlgName{} (\textsc{Ours})}} 
& 256 & 61.90 & 39.66 & 41.18 & 55.65 & 85.82 & 33.78 & 86.25 & 75.89 & 25.20 & 65.66 & 29.63 \\
& 512 & 62.97 & 49.79 & 43.14 & 57.91 & 86.57 & 52.70 & 83.70 & 85.71 & 27.40 & 60.61 & 30.86 \\
& 1024 & 64.70 & 52.74 & 49.02 & 72.60 & 85.07 & 41.89 & 81.83 & 89.29 & 29.20 & 64.65 & 31.48 \\
\midrule

\multirow{3}{*}{\textbf{\textsc{EmbedLLM}}} 
& 256 & 58.80 & 58.23 & 41.18 & 72.03 & 79.85 & 35.14 & 75.55 & 88.39 & 28.20 & 19.19 & 30.25 \\
& 512 & 59.47 & 59.49 & 45.10 & 74.29 & 77.61 & 43.24 & 76.83 & 89.29 & 28.80 & 12.12 & 29.63 \\
& 1024 & 59.60 & 58.65 & 50.98 & 72.88 & 79.10 & 47.30 & 76.83 & 89.29 & 23.80 & 25.25 & 30.86 \\

\midrule

\multirow{3}{*}{\textbf{\textsc{EmbedLLM+}}} 
& 256 & 59.17 & 49.37 & 35.29 & 62.99 & 62.69 & 51.35 & 77.08 & 89.29 & 24.20 & 54.55 & 35.80 \\
& 512 & 61.57 & 57.81 & 45.10 & 73.45 & 81.34 & 41.89 & 78.27 & 89.29 & 29.00 & 33.84 & 32.72 \\
& 1024 & 59.73 & 57.38 & 45.10 & 71.75 & 79.85 & 32.43 & 78.69 & 89.29 & 26.20 & 21.72 & 29.01 \\
\midrule

\multirow{3}{*}{\textbf{\textsc{IRT-Net}}} 
& 256 & 59.57 & 59.49 & 39.22 & 72.60 & 81.34 & 43.24 & 76.91 & 90.18 & 28.60 & 14.65 & 30.25 \\
& 512 & 60.17 & 58.65 & 49.02 & 72.60 & 80.60 & 44.59 & 77.25 & 83.04 & 28.40 & 23.23 & 32.10 \\
& 1024 & 63.37 & 57.81 & 52.94 & 71.19 & 79.85 & 50.00 & 82.17 & 88.39 & 27.60 & 43.94 & 30.25 \\
\midrule

& & \multicolumn{11}{c}{\textbf{Embedding Free Routers}} \\
\addlinespace[2pt]
\midrule

\multirow{3}{*}{\textbf{\textsc{kNN}}} 
& 256 & 60.50 & 59.92 & 47.06 & 75.42 & 85.07 & 41.89 & 78.27 & 89.29 & 29.00 & 10.10 & 30.86 \\
& 512 & 61.37 & 57.38 & 45.10 & 68.64 & 85.82 & 48.65 & 80.48 & 89.29 & 26.80 & 29.80 & 29.01 \\
& 1024 & 62.90 & 57.81 & 45.10 & 72.60 & 86.57 & 55.41 & 82.09 & 89.29 & 21.60 & 39.39 & 37.04 \\
\midrule

\multirow{3}{*}{\textbf{\textsc{KMeans}}} 
& 256 & 61.33 & 58.65 & 49.02 & 62.99 & 86.57 & 52.70 & 84.72 & 89.29 & 25.80 & 10.10 & 31.48 \\
& 512 & 64.47 & 57.81 & 52.94 & 75.42 & 87.31 & 47.30 & 81.32 & 89.29 & 23.60 & 57.58 & 37.65 \\
& 1024 & 66.20 & 56.12 & 49.02 & 74.86 & 87.31 & 52.70 & 85.14 & 89.29 & 24.80 & 60.10 & 37.65 \\

\bottomrule
\end{tabular}

\end{table*}

\begin{table*}[htbp]
\centering
\caption{Overall and per-dataset correctness prediction accuracy, across approaches and training-set sizes. We compare against model embedding generation approaches EmbedLLM \cite{ICLR2025_embedllm}, EmbedLLM+ (with more hidden layers than original \cite{ICLR2025_embedllm}) and IRT-Net \cite{chen2025learningcompactrepresentationsllm_irtnet}. We also present performance of nonparametric routers ($K$-Means and $k$NN) as strong baselines on correctness prediction of models on queries. Across all queries, the overall correctness prediction accuracy is similar for all presented approaches. 
}
\label{tab:correctness_pred_by_dataset_full}
\scriptsize
\setlength{\tabcolsep}{2.5pt}      
\renewcommand{\arraystretch}{1.15} 

\begin{tabular}{@{}l c c *{10}{c}@{}}
\toprule
\multirow{2}{*}{\textbf{Approach}} &
\multirow{2}{*}{\textbf{Train samples}} &
\multicolumn{11}{c}{\textbf{Correctness Prediction Accuracy (\%)}} \\
\cmidrule(lr){3-13}
& & \textbf{Overall} & MathQA & LogiQA & MedQA & PIQA & TruthQA & MMLU & GSM8k & GPQA & ASDiv & SoQA \\
\midrule

\multirow{3}{*}{\textbf{\AlgName{} (\textsc{Ours})}} 
& 256 & 68.31 & 65.38 & 64.88 & 59.36 & 72.21 & 69.12 & 62.99 & 66.18 & 79.17 & 96.46 & 61.76 \\
& 512 & 68.33 & 64.09 & 59.84 & 60.03 & 65.78 & 69.12 & 64.64 & 71.75 & 79.18 & 96.05 & 54.08 \\
& 1024 & 70.03 & 65.22 & 67.75 & 60.89 & 73.03 & 67.43 & 65.39 & 72.85 & 79.42 & 96.50 & 66.83 \\
\midrule

\multirow{3}{*}{\textbf{\textsc{EmbedLLM}}} 
& 256 & 67.33 & 65.95 & 55.58 & 60.39 & 57.24 & 40.08 & 64.01 & 62.32 & 79.03 & 96.14 & 65.24 \\
& 512 & 68.12 & 65.75 & 55.92 & 58.78 & 76.86 & 39.18 & 64.80 & 65.51 & 78.73 & 95.68 & 61.40 \\
& 1024 & 69.47 & 65.50 & 61.38 & 60.45 & 76.61 & 60.56 & 65.62 & 70.07 & 79.02 & 94.95 & 62.67 \\
\midrule

\multirow{3}{*}{\textbf{\textsc{EmbedLLM+}}} 
& 256 & 66.14 & 65.09 & 53.15 & 58.83 & 47.88 & 38.04 & 62.77 & 65.95 & 78.69 & 95.28 & 66.01 \\
& 512 & 68.09 & 64.29 & 57.63 & 58.89 & 68.97 & 51.35 & 65.19 & 68.53 & 78.08 & 94.56 & 61.63 \\
& 1024 & 69.33 & 65.22 & 63.25 & 60.19 & 74.41 & 56.38 & 65.55 & 70.94 & 78.47 & 95.72 & 64.84 \\
\midrule

\multirow{3}{*}{\textbf{\textsc{IRT-Net}}} 
& 256 & 67.38 & 65.74 & 59.89 & 60.57 & 57.39 & 47.09 & 63.57 & 64.23 & 78.87 & 95.18 & 65.08 \\
& 512 & 69.07 & 65.12 & 62.31 & 59.69 & 75.69 & 57.89 & 65.20 & 68.87 & 78.59 & 95.32 & 64.00 \\
& 1024 & 70.12 & 65.85 & 65.07 & 61.06 & 77.43 & 65.69 & 65.83 & 70.13 & 79.02 & 96.22 & 65.51 \\
\midrule

& & \multicolumn{11}{c}{\textbf{Embedding Free Routers}} \\
\addlinespace[2pt]
\midrule

\multirow{3}{*}{\textbf{\textsc{kNN}}} 
& 256 & 60.60 & 65.25 & 65.35 & 59.73 & 26.33 & 66.31 & 52.08 & 62.65 & 74.43 & 88.25 & 63.94 \\
& 512 & 67.90 & 60.47 & 55.02 & 60.10 & 78.38 & 50.86 & 63.82 & 73.92 & 78.21 & 89.84 & 65.90 \\
& 1024 & 69.27 & 62.81 & 59.58 & 60.55 & 76.32 & 65.82 & 64.54 & 74.00 & 78.93 & 94.19 & 67.48 \\
\midrule

\multirow{3}{*}{\textbf{\textsc{KMeans}}} 
& 256 & 66.34 & 66.15 & 55.74 & 59.39 & 65.51 & 35.34 & 61.33 & 58.45 & 78.76 & 95.57 & 67.74 \\
& 512 & 68.32 & 66.09 & 51.98 & 61.29 & 79.32 & 44.63 & 63.25 & 73.71 & 76.37 & 96.69 & 67.47 \\
& 1024 & 69.57 & 65.59 & 55.57 & 61.20 & 79.64 & 62.90 & 64.16 & 74.04 & 78.67 & 96.74 & 67.85 \\

\bottomrule
\end{tabular}

\end{table*}

\newpage
~
\newpage

\section{Ablation on choice of Sentence Encoder for Query Representation} \label{app:encoder_ablation}

\begin{table*}[h]
\centering
\caption{Overall and per-dataset Routing Accuracy, across query sentence encoder choices.}
\label{tab:routing_acc_by_dataset_full_encoder_ablation}
\scriptsize
\setlength{\tabcolsep}{2.5pt}      
\renewcommand{\arraystretch}{1.15}

\begin{tabular}{@{}l c c *{10}{c}@{}}
\toprule
\multirow{2}{*}{\textbf{Sentence Encoder}} &
\multirow{2}{*}{\textbf{Train samples}} &
\multicolumn{11}{c}{\textbf{Routing Accuracy (\%)}} \\
\cmidrule(lr){3-13}
& & \textbf{Overall} & MathQA & LogiQA & MedQA & PIQA & TruthQA & MMLU & GSM8k & GPQA & ASDiv & SoQA \\
\midrule

\multirow{3}{*}{\shortstack[l]{\texttt{all\_mpnet\_base\_v2}\\\cite{song2020mpnetmaskedpermutedpretraining}}}
& 256 & 61.90 & 39.66 & 41.18 & 55.65 & 85.82 & 33.78 & 86.25 & 75.89 & 25.20 & 65.66 & 29.63 \\
& 512 & 62.97 & 49.79 & 43.14 & 57.91 & 86.57 & 52.70 & 83.70 & 85.71 & 27.40 & 60.61 & 30.86 \\
& 1024 & 64.70 & 52.74 & 49.02 & 72.60 & 85.07 & 41.89 & 81.83 & 89.29 & 29.20 & 64.65 & 31.48 \\
\midrule

\multirow{3}{*}{\shortstack[l]{\texttt{paraphrase\_albert\_small\_v2}\\\cite{Lan2020ALBERT}}}
& 256 & 59.00 & 43.88 & 45.10 & 57.34 & 86.57 & 25.68 & 85.40 & 80.36 & 26.00 & 14.65 & 30.86 \\
& 512 & 61.40 & 58.23 & 49.02 & 64.97 & 85.82 & 41.89 & 84.30 & 88.39 & 28.40 & 9.60 & 30.86 \\
& 1024 & 60.97 & 59.07 & 52.94 & 69.77 & 86.57 & 41.89 & 80.98 & 88.39 & 29.00 & 10.10 & 30.86 \\

\midrule

\multirow{3}{*}{\shortstack[l]{\texttt{all\_minilm\_l6\_v2}\\\cite{minilm_sentence_encoder}}}
& 256 & 61.60 & 59.92 & 49.02 & 70.90 & 87.31 & 41.89 & 82.17 & 89.29 & 28.80 & 10.10 & 30.86 \\
& 512 & 63.20 & 83.45 & 27.00 & 67.80 & 51.48 & 49.49 & 30.86 & 82.84 & 89.29 & 41.89 & 50.98 \\
& 1024 & 62.27 & 83.28 & 23.60 & 61.58 & 54.43 & 55.56 & 30.86 & 87.31 & 88.39 & 32.43 & 43.14 \\
\midrule

\multirow{3}{*}{\shortstack[l]{\texttt{all\_distilroberta\_v1}\\\cite{sanh2020distilbertdistilledversionbert}}}
& 256 & 61.83 & 82.09 & 24.80 & 58.76 & 45.15 & 66.16 & 30.25 & 85.07 & 82.14 & 52.70 & 47.06 \\
& 512 & 62.60 & 82.09 & 25.80 & 62.71 & 50.21 & 61.62 & 27.16 & 86.57 & 85.71 & 48.65 & 52.94 \\
& 1024 & 63.33 & 82.26 & 24.20 & 66.10 & 54.01 & 66.16 & 31.48 & 82.84 & 89.29 & 41.89 & 47.06 \\

\bottomrule
\end{tabular}

\end{table*}

\begin{table*}[h]
\centering
\caption{Overall and per-dataset Correctness Prediction Accuracy, across query sentence encoder choices.}
\label{tab:correctness_pred_by_dataset_full_encoder_ablation}
\scriptsize
\setlength{\tabcolsep}{2.5pt}      
\renewcommand{\arraystretch}{1.15} 

\begin{tabular}{@{}l c c *{10}{c}@{}}
\toprule
\multirow{2}{*}{\textbf{Sentence Encoder}} &
\multirow{2}{*}{\textbf{Train samples}} &
\multicolumn{11}{c}{\textbf{Correctness Prediction Accuracy (\%)}} \\
\cmidrule(lr){3-13}
& & \textbf{Overall} & MathQA & LogiQA & MedQA & PIQA & TruthQA & MMLU & GSM8k & GPQA & ASDiv & SoQA \\
\midrule

\multirow{3}{*}{\shortstack[l]{\texttt{all\_mpnet\_base\_v2}\\\cite{song2020mpnetmaskedpermutedpretraining}}}
& 256 & 68.31 & 65.38 & 64.88 & 59.36 & 72.21 & 69.12 & 62.99 & 66.18 & 79.17 & 96.46 & 61.76 \\
& 512 & 68.33 & 64.09 & 59.84 & 60.03 & 65.78 & 69.12 & 64.64 & 71.75 & 79.18 & 96.05 & 54.08 \\
& 1024 & 70.03 & 65.22 & 67.75 & 60.89 & 73.03 & 67.43 & 65.39 & 72.85 & 79.42 & 96.50 & 66.83 \\
\midrule

\multirow{3}{*}{\shortstack[l]{\texttt{paraphrase\_albert\_small\_v2}\\\cite{Lan2020ALBERT}}}
& 256 & 59.42 & 66.33 & 67.31 & 57.29 & 19.50 & 69.12 & 46.38 & 57.75 & 79.73 & 96.55 & 68.07 \\
& 512 & 66.61 & 64.38 & 58.40 & 59.08 & 59.95 & 69.12 & 62.54 & 57.28 & 79.63 & 96.20 & 53.04 \\
& 1024 & 67.69 & 64.04 & 65.37 & 56.58 & 73.69 & 64.56 & 64.18 & 57.14 & 78.56 & 94.53 & 60.92 \\
\midrule

\multirow{3}{*}{\shortstack[l]{\texttt{all\_minilm\_l6\_v2}\\\cite{minilm_sentence_encoder}}}
& 256 & 66.12 & 65.07 & 63.03 & 58.51 & 54.56 & 67.86 & 61.89 & 58.69 & 76.45 & 95.10 & 62.51 \\
& 512 & 68.14 & 63.85 & 78.50 & 59.62 & 65.58 & 96.26 & 61.12 & 62.49 & 70.30 & 68.64 & 59.05 \\
& 1024 & 69.28 & 64.47 & 79.26 & 60.67 & 65.44 & 96.11 & 66.05 & 69.62 & 71.35 & 66.19 & 65.06 \\
\midrule

\multirow{3}{*}{\shortstack[l]{\texttt{all\_distilroberta\_v1}\\\cite{sanh2020distilbertdistilledversionbert}}} 
& 256 & 67.50 & 61.78 & 78.35 & 58.92 & 63.43 & 95.89 & 58.21 & 71.33 & 72.08 & 69.12 & 68.79 \\
& 512 & 68.17 & 64.16 & 78.85 & 59.61 & 65.99 & 95.86 & 55.52 & 64.03 & 71.07 & 68.96 & 61.33 \\
& 1024 & 70.34 & 65.85 & 79.11 & 61.24 & 65.98 & 96.75 & 67.27 & 74.31 & 73.40 & 67.98 & 64.92 \\

\bottomrule
\end{tabular}

\end{table*}

In this section, we present correctness prediction accuracy and routing accuracy of \AlgName{} through different choices of sentence encoders for generating query representations. In our main analysis, we select a popular sentence encoder \texttt{all\_mpnet\_base\_v2} \cite{song2020mpnetmaskedpermutedpretraining} which produces query representations $\in\mathbb{R}^{768}$. 

Here, we additionally present results for \texttt{paraphrase\_albert\_small\_v2} \cite{Lan2020ALBERT} ($\in\mathbb{R}^{768}$), \texttt{all\_minilm\_l6\_v2} \cite{minilm_sentence_encoder} ($\in\mathbb{R}^{384}$), and \texttt{all\_distilroberta\_v1} \cite{sanh2020distilbertdistilledversionbert} ($\in\mathbb{R}^{768}$). Our model architecture is uniform across all sentence encoders (except for different input dimensions across choices of sentence encoders). We present routing accuracy in \cref{tab:routing_acc_by_dataset_full_encoder_ablation} and correctness prediction accuracy in \cref{tab:correctness_pred_by_dataset_full_encoder_ablation}. 

We observe that overall routing accuracy and correctness prediction accuracy is comparable across sentence encoders, but individual dataset-wise performance varies. We attribute this to varying query representation capabilities across tasks for different sentence encoders. 

\newpage

\section{Ablation on Embedding Dimension} \label{app:embedding_dim_ablation}

Here, we present experimental results on our encoder-decoder approach $(F_\theta,G_\psi)$ with varying model embedding dimension (and corresponding internal dimensions as well). \Cref{tab:routing_acc_by_dataset_full_embed_dim_ablation} denotes routing accuracy, and \cref{tab:correctness_pred_by_dataset_full_emb_dim_ablation} denotes correctness prediction accuracy for embedding dimensions $\in\{16,32,64,128,256,512\}$. We find that model embedding dimensions $128/256$ show superior routing accuracy and correctness prediction accuracy, whereas smaller embedding dimensions are not informative enough and larger embedding dimensions increase training complexity, eventually affecting numerical performance.

\begin{table*}[ht]
\centering
\caption{Overall and per-dataset Routing Accuracy, across model embedding dimension choices.}
\label{tab:routing_acc_by_dataset_full_embed_dim_ablation}
\scriptsize
\setlength{\tabcolsep}{2.5pt}      
\renewcommand{\arraystretch}{1}

\begin{tabular}{@{}l c c *{10}{c}@{}}
\toprule
\multirow{2}{*}{\textbf{Model Embedding Dimension}} &
\multirow{2}{*}{\textbf{Train samples}} &
\multicolumn{11}{c}{\textbf{Routing Accuracy (\%)}} \\
\cmidrule(lr){3-13}
& & \textbf{Overall} & MathQA & LogiQA & MedQA & PIQA & TruthQA & MMLU & GSM8k & GPQA & ASDiv & SoQA \\
\midrule

\multirow{3}{*}{16}
& 256 & 60.97 & 41.35 & 43.14 & 55.08 & 83.58 & 50.00 & 82.09 & 79.46 & 24.40 & 68.69 & 31.48 \\
& 512 & 62.93 & 45.57 & 43.14 & 66.10 & 84.33 & 45.95 & 82.43 & 89.29 & 27.60 & 61.11 & 29.01 \\
& 1024 & 63.87 & 51.48 & 52.94 & 71.19 & 82.84 & 50.00 & 79.97 & 90.18 & 29.00 & 65.66 & 30.25 \\
\midrule

\multirow{3}{*}{32}
& 256 & 62.13 & 41.35 & 39.22 & 64.69 & 85.07 & 25.68 & 84.13 & 81.25 & 26.40 & 60.61 & 30.86 \\
& 512 & 62.17 & 50.21 & 50.98 & 66.10 & 83.58 & 50.00 & 80.22 & 87.50 & 26.00 & 58.59 & 29.63 \\
& 1024 & 62.90 & 50.21 & 47.06 & 66.67 & 79.85 & 41.89 & 80.48 & 89.29 & 28.40 & 66.16 & 30.25 \\

\midrule

\multirow{3}{*}{64}
& 256 & 61.77 & 32.49 & 41.18 & 57.06 & 84.33 & 45.95 & 85.40 & 82.14 & 25.20 & 66.67 & 30.86 \\
 & 512 & 61.60 & 48.10 & 47.06 & 57.91 & 85.07 & 41.89 & 81.24 & 82.14 & 27.60 & 62.63 & 30.25 \\
& 1024 & 62.87 & 52.32 & 45.10 & 61.30 & 86.57 & 44.59 & 83.02 & 78.57 & 24.40 & 64.65 & 35.19 \\
\midrule

\multirow{3}{*}{128 (default)}
& 256 & 61.90 & 39.66 & 41.18 & 55.65 & 85.82 & 33.78 & 86.25 & 75.89 & 25.20 & 65.66 & 29.63 \\
& 512 & 62.97 & 49.79 & 43.14 & 57.91 & 86.57 & 52.70 & 83.70 & 85.71 & 27.40 & 60.61 & 30.86 \\
& 1024 & 64.70 & 52.74 & 49.02 & 72.60 & 85.07 & 41.89 & 81.83 & 89.29 & 29.20 & 64.65 & 31.48 \\
\midrule

\multirow{3}{*}{256}
& 256 & 61.87 & 35.86 & 43.14 & 61.86 & 85.07 & 25.68 & 84.30 & 78.57 & 27.20 & 65.66 & 30.86 \\
& 512 & 63.27 & 48.95 & 45.10 & 61.86 & 84.33 & 51.35 & 83.45 & 87.50 & 28.20 & 61.11 & 28.40 \\
& 1024 & 64.43 & 53.59 & 52.94 & 72.88 & 87.31 & 41.89 & 81.66 & 81.25 & 27.20 & 68.18 & 30.25 \\
\midrule

\multirow{3}{*}{512}
& 256 & 62.10 & 37.97 & 39.22 & 54.80 & 85.82 & 50.00 & 85.57 & 78.57 & 26.80 & 65.15 & 29.63 \\
& 512 & 64.07 & 54.01 & 47.06 & 64.41 & 84.33 & 52.70 & 83.36 & 88.39 & 27.60 & 60.61 & 31.48 \\
& 1024 & 62.70 & 53.59 & 35.29 & 64.12 & 86.57 & 54.05 & 80.73 & 78.57 & 26.00 & 65.66 & 33.33 \\

\bottomrule
\end{tabular}

\end{table*}

\begin{table*}[ht]
\centering
\caption{Overall and per-dataset Correctness Prediction Accuracy, across model embedding dimension choices.}
\label{tab:correctness_pred_by_dataset_full_emb_dim_ablation}
\scriptsize
\setlength{\tabcolsep}{2.5pt}     
\renewcommand{\arraystretch}{1.}

\begin{tabular}{@{}l c c *{10}{c}@{}}
\toprule
\multirow{2}{*}{\textbf{Model Embedding Dimension}} &
\multirow{2}{*}{\textbf{Train samples}} &
\multicolumn{11}{c}{\textbf{Correctness Prediction Accuracy (\%)}} \\
\cmidrule(lr){3-13}
& & \textbf{Overall} & MathQA & LogiQA & MedQA & PIQA & TruthQA & MMLU & GSM8k & GPQA & ASDiv & SoQA \\
\midrule

\multirow{3}{*}{16}
& 256 & 67.94 & 64.38 & 65.72 & 58.63 & 69.48 & 69.05 & 62.76 & 67.71 & 78.93 & 96.48 & 61.54 \\
& 512 & 68.32 & 64.58 & 59.84 & 60.33 & 64.95 & 69.08 & 64.06 & 71.29 & 79.24 & 96.22 & 57.45 \\
& 1024 & 69.97 & 65.75 & 67.79 & 61.02 & 73.38 & 69.12 & 65.24 & 71.26 & 79.11 & 96.40 & 66.95 \\
\midrule

\multirow{3}{*}{32}
& 256 & 67.75 & 64.73 & 65.20 & 58.52 & 72.33 & 69.12 & 62.19 & 68.67 & 78.26 & 96.30 & 61.09 \\
& 512 & 68.47 & 64.66 & 60.08 & 60.25 & 66.90 & 69.06 & 64.42 & 70.97 & 79.03 & 96.24 & 56.88 \\
& 1024 & 69.43 & 65.68 & 68.14 & 60.62 & 70.80 & 66.37 & 64.41 & 71.60 & 79.24 & 96.44 & 66.51 \\

\midrule

\multirow{3}{*}{64}
& 256 & 67.82 & 65.53 & 65.51 & 58.75 & 68.38 & 69.12 & 61.72 & 69.15 & 79.51 & 96.71 & 62.64 \\
& 512 & 68.53 & 65.06 & 61.50 & 60.05 & 66.44 & 69.11 & 64.24 & 70.75 & 79.15 & 96.36 & 58.69 \\
& 1024 & 69.73 & 65.14 & 67.19 & 60.67 & 73.63 & 66.05 & 65.37 & 71.97 & 78.94 & 96.66 & 64.25 \\
\midrule

\multirow{3}{*}{128 (default)}
& 256 & 68.31 & 65.38 & 64.88 & 59.36 & 72.21 & 69.12 & 62.99 & 66.18 & 79.17 & 96.46 & 61.76 \\
& 512 & 68.33 & 64.09 & 59.84 & 60.03 & 65.78 & 69.12 & 64.64 & 71.75 & 79.18 & 96.05 & 54.08 \\
& 1024 & 70.03 & 65.22 & 67.75 & 60.89 & 73.03 & 67.43 & 65.39 & 72.85 & 79.42 & 96.50 & 66.83 \\
\midrule

\multirow{3}{*}{256}
& 256 & 68.42 & 65.14 & 64.79 & 59.82 & 75.03 & 69.12 & 63.22 & 67.24 & 78.87 & 96.44 & 59.60 \\
& 512 & 68.36 & 63.76 & 60.59 & 60.06 & 69.71 & 68.85 & 64.54 & 70.37 & 78.88 & 96.08 & 54.31 \\
& 1024 & 70.34 & 65.39 & 67.86 & 61.64 & 73.99 & 66.86 & 65.81 & 72.97 & 79.21 & 96.68 & 67.39 \\
\midrule

\multirow{3}{*}{512}
& 256 & 67.93 & 65.22 & 65.86 & 59.35 & 72.06 & 69.12 & 62.31 & 67.85 & 79.05 & 96.14 & 59.53 \\
& 512 & 68.35 & 64.57 & 60.82 & 60.20 & 63.97 & 69.12 & 64.84 & 71.01 & 78.96 & 96.20 & 54.20 \\
& 1024 & 69.89 & 65.53 & 67.35 & 60.74 & 70.78 & 68.98 & 65.43 & 71.67 & 79.13 & 96.80 & 66.50 \\

\bottomrule
\end{tabular}

\end{table*}

\newpage

\section{Additional Utility of Model Embeddings} \label{app:additional_utility_embeddings}

\subsection{Retrieving models that match desired task profiles via hypothetical embeddings.}\label{subapp:hypothetical_embeddings}
We next test whether embeddings preserve task-level performance structure rather than overfitting to query identity. For each model $m\in \mathcal{M}$, we compute its empirical task-wise accuracy profile across a set of tasks. We then construct a \emph{hypothetical} evaluation set by sampling queries and assigning synthetic correctness labels so that the resulting task-wise accuracies match (in expectation) the target profile of $m$. Feeding this synthetic set through the encoder produces a hypothetical embedding $\tilde{z}_{m}$, which we then compare against the library of real model embeddings $\{z_m\}_{m\in\mathcal{M}}$ via nearest-neighbor search.

\begin{figure}[htbp]
    \centering

    \begin{minipage}[t]{0.49\linewidth}
        \centering
        \includegraphics[width=0.8\linewidth]{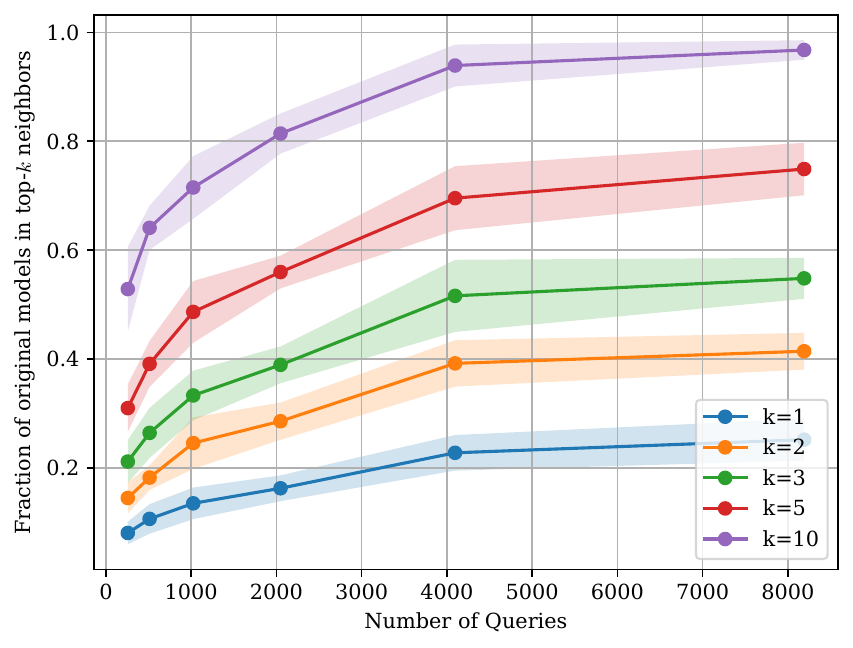}
        \captionof{figure}{We obtain model embeddings based on random assignment of correctness labels to queries to match dataset-wise accuracy scores of actual LLMs in the pool. Then we find the proportion of original model embeddings that are among the ($1,2,3,5,10$) nearest neighbors of the new generated embedding. A high fraction of original embeddings present in top-$k$ neighbors of the new generated embeddings denotes that \AlgName{} captures task level behavior in model embedding geometry.}
        \label{fig:hypothetical_model_embedding}
    \end{minipage}\hfill
    \begin{minipage}[t]{0.49\linewidth}
        \centering
        \includegraphics[width=\linewidth]{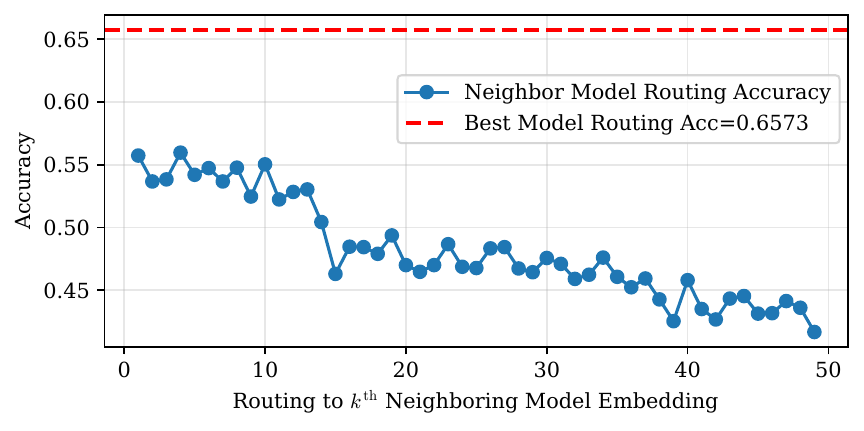}
        \captionof{figure}{\textbf{Resilient Routing to Fallback Models.} Selected models for query routing might not always be available due to congestion, node failure etc. Routing to next closest model embedding provides a fallback mechanism, while maintaining up to $85\%$ of overall routing performance. Performance gracefully decays with chosen rank of neighboring model to route queries to. }
        \label{fig:fallback_routing}
    \end{minipage}

\end{figure}

We report recall@k: the fraction of models for which the true model embedding $z_m$ appears among the top-$k$ nearest real embeddings to $\tilde{z}_{m}$. Averaged over all models, recall@10 is almost $97\%$ for evaluation sets of size $8192$ queries, depicting that model embeddings capture high level task performance (\cref{fig:hypothetical_model_embedding}). These results indicate that the embedding generator maps evaluation sets to representations that reflect higher-level behavioral profiles: even when correctness labels are randomized at the query level subject to matching task-level rates, the resulting embedding is often close to the intended model.

\subsection{Resilient fallback routing under model unavailability.} \label{subapp:resilient_routing}

In deployed query routing systems, the language model selected by a router may become temporarily unreachable (e.g., due to overload, failures, or policy constraints). When routing decisions are produced by a black-box router and per-query correctness probabilities / suitability scores for models are not readily available, a natural fallback is to substitute the router selected model with a nearby model in embedding space. To study the validity of the fallback routing approach, for each test query $x$ we take the routed model $\widehat{m}(x)$ and simulate its unavailability, then generate a  response using its $k^{\text{th}}$ nearest neighbor by embedding distance (closest, second-closest, etc.).  \Cref{fig:fallback_routing} reports routing accuracy versus neighbor rank: the closest-neighbor fallback retains up to $\mathbf{85\%}$ of the original routing accuracy, with performance decreasing smoothly as $k$ increases. This supports embedding proximity as a practical mechanism for resilient fallback under model outages.

\newpage
\subsection{Convergence relative to inter-model separation, and implications to security fingerprinting.}\label{subapp:embedding_convergence_security}
\begin{figure}[htbp]
    \centering
    \includegraphics[width=0.8\linewidth]{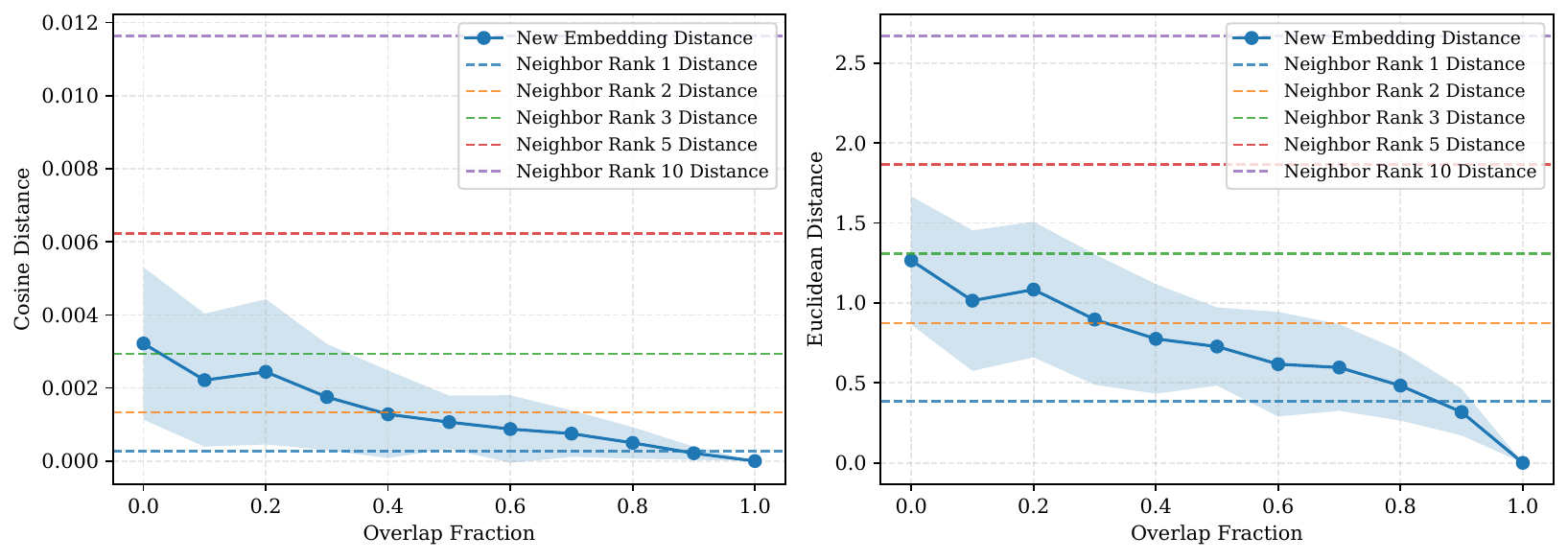}
    \caption{Distance between new generated model embedding to reference embedding (for CodeLlama-13b-hf model) w.r.t. changing the overlap fraction of queries in evaluation set used for generating model embeddings. Horizontal dashed lines denote distance of chosen model's reference embedding to other models' embeddings.}
    \label{fig:overlap_convergence}
\end{figure}

\begin{figure}[htbp]
    \centering
    \includegraphics[width=0.8\linewidth]{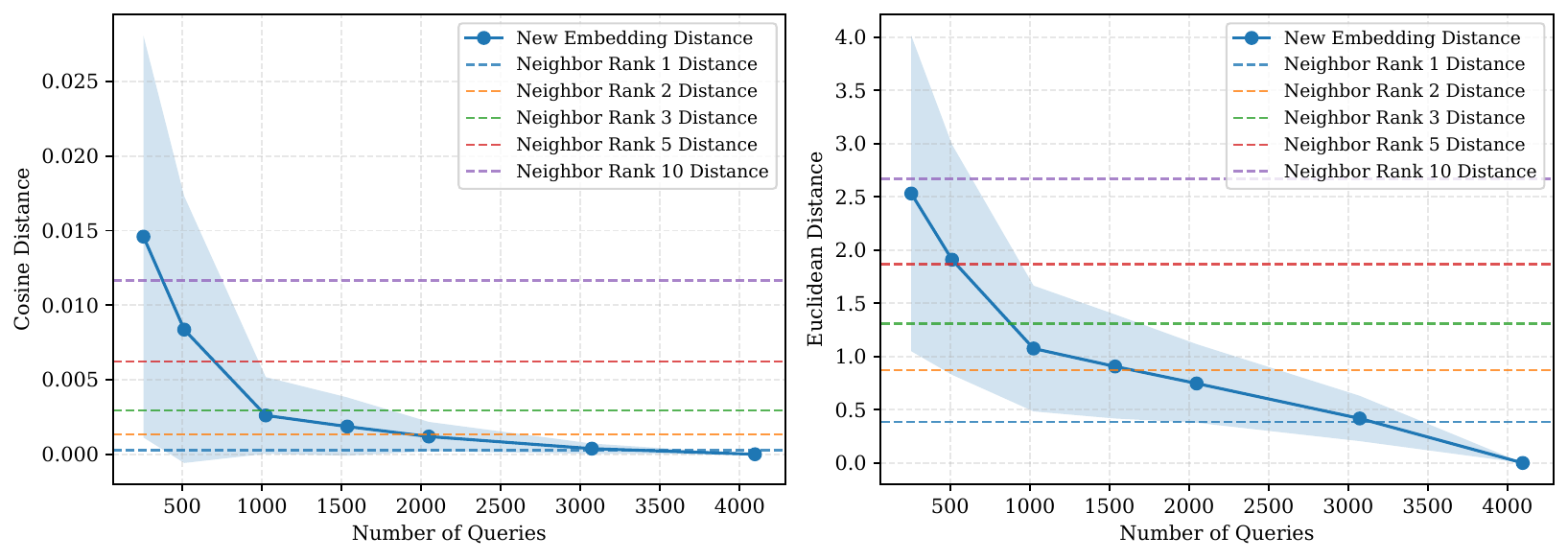}
    \caption{Distance between new generated model embedding to reference embedding (for CodeLlama-13b-hf model) w.r.t. changing the number of queries in evaluation set used for generating model embeddings. Horizontal dashed lines denote distance of chosen model's reference embedding to other models' embeddings.}
    \label{fig:anchor_size_convergence}
\end{figure}

To quantify and contextualize the model embedding drift w.r.t. varying the evaluation set overlap and size, we plot the distance to the originally generated reference embedding $\|z_m - z_m^{\mathrm{ref}}\|$ and compare them with embedding distances of reference embedding of chosen model to reference embeddings of other distinct models. \Cref{fig:overlap_convergence} shows that the new model embeddings generated are close to the reference embeddings (as compared to distances to other reference model embeddings), and the distance/dissimilarity decays as the overlap fraction (or the number of evaluation queries) increases. A similar trend is also observed for subsampling a smaller number of queries in the evaluation set for generating model embeddings, as presented in \cref{fig:anchor_size_convergence}. 

Relative robustness of generated model embeddings to the choice of queries demonstrates possible applications in security fingerprinting of language models, where very close embeddings between two models can indicate the same language model hidden behind different APIs or endpoints.

\newpage
\subsection{Model portfolio selection using embedding geometry}\label{subapp:model_portfolio}

\subsubsection{Setup and portfolio routing accuracy}
Let $\mathcal{M}=\{1,\dots,M\}$ denote the full model pool and let $z_m\in\mathbb{R}^d$ be the learned embedding for model $m$.
Given a query $x$, the router assigns a correctness probability $\widehat{p}^{(m)}(x)$ to each model $m$.
For a selected portfolio $\mathcal{M}_\textrm{sel}\subseteq\mathcal{M}$, routing chooses
\begin{equation}
\widehat m_\textrm{sel}(x) \;=\; \argmax_{m\in \mathcal{M}_\textrm{sel}} \widehat{p}^{(m)}(x).
\end{equation}
Let $y^{(m)}(x)\in\{0,1\}$ indicate whether model $m$ answers $x$ correctly (used only for evaluation).
We define \emph{portfolio routing accuracy} as
\begin{equation}
\mathrm{Acc}(\mathcal{M}_\textrm{sel}) \;=\; \mathbb{E}_{x}\big[y^{(\widehat m_\textrm{sel})}(x)\big],
\end{equation}
which we estimate on held-out data as routing accuracy after restricting to the models in $\mathcal{M}_\textrm{sel}$.
Our goal is to select $\mathcal{M}_\textrm{sel}$ using \emph{only} embedding geometry $\{z_m\}$ under two different deployment constraints:
\begin{enumerate}
    \item a \emph{count} constraint $|\mathcal{M}_\textrm{sel}|\leq k$, with $k\in\{1,2,3,\cdots\}$
    \item a \emph{parameter budget} constraint $\sum_{m\in \mathcal{M}_\textrm{sel}} c_m \le \mathcal{C}$, where $c_m>0$ denotes total number of parameters in model $m$.
\end{enumerate}

\paragraph{Similarity between model embeddings through Kernels.}

Let $K(z_i,z_j)\in[0,1]$ be a similarity kernel in the model embedding space (e.g., an RBF kernel),
where larger values indicate close model embeddings and hence similar model behaviors. Concretely, we define
\begin{equation}
K(z_i,z_j) \;=\; \exp\!\Big(\! -\tfrac{d_z(z_i,z_j)^2}{\sigma^2}\Big),
\label{eq:rbf_kernel}
\end{equation}
with $d_z(\cdot,\cdot)$ being a distance/dissimilarity measure (e.g., cosine distance) and bandwidth $\sigma>0$ (which we set as the median pairwise distance between model embeddings).

\subsubsection{Model Count Constrained Selection}\label{subsubapp:count_constrained_portfolio_selection}

To choose a subset of models such that collective abilities of the pool is maintained, we select model embeddings spread across the embedding space via the kernelized $k$-center objective \cite{GONZALEZ1985293_k_centers}:
\begin{equation}
\max_{\mathcal{M}_\textrm{sel} \subseteq \mathcal{M}:\ |\mathcal{M}_\textrm{sel}|\le k}\;\; 
\min_{i\in \mathcal{M}}\;\max_{s\in \mathcal{M}_\textrm{sel}} K(z_i,z_s),
\label{eq:kcenter_obj_kernel}
\end{equation}
which seeks to choose $k$ model embeddings such that every model in the full pool has \emph{high similarity} to at least one chosen model in the subset (i.e., the worst-covered model is as well-covered as possible).

We additionally study the kernelized $k$-medoids objective \cite{KaufmanRousseeuw1990PAM_k_medoids} which maximizes the sum similarity of all model embeddings to the most similar among the chosen subset of models closest :
\begin{equation}
\max_{\mathcal{M}_\textrm{sel} \subseteq \mathcal{M}:\ |\mathcal{M}_\textrm{sel}|\le k}\;\;
\sum_{i\in \mathcal{M}} \max_{s\in \mathcal{M}_\textrm{sel}} K(z_i,z_s),
\label{eq:kmedoids_obj_kernel}
\end{equation}
which selects $k$ embeddings so that the \emph{total coverage} (best similarity per model, summed over the pool) is maximized.

Both aforementioned approaches depend only on model embedding geometry and do not utilize per-query correctness.

\paragraph{\texttt{k-center} (farthest-first in similarity space) \cite{GONZALEZ1985293_k_centers}.}
We apply a similarity-space analogue of farthest-first sampling. We initialize $\mathcal{M}_\textrm{sel}=\{m_0\}$ using the \emph{global kernel medoid}
\begin{equation}
m_0 \;=\; \argmax_{m\in\mathcal{M}}\;\frac{1}{|\mathcal{M}|}\sum_{i\in\mathcal{M}} K(z_i,z_m),
\end{equation}
then iteratively add the \emph{least-covered} model (the one whose best similarity to the current set is smallest):
\begin{equation}
m^\star \;=\; \argmin_{m\in \mathcal{M}\setminus \mathcal{M}_\textrm{sel}}\;\; \max_{s\in \mathcal{M}_\textrm{sel}} K(z_m,z_s),
\qquad \mathcal{M}_\textrm{sel} \leftarrow \mathcal{M}_\textrm{sel} \cup \{m^\star\}.
\label{eq:farthest_first_kernel}
\end{equation}
Intuitively, this repeatedly adds a representative for the region of embedding space that is poorly covered at each step (lowest similarity to models in the selected set $\mathcal{M}_\textrm{sel}$).

\paragraph{\texttt{k-medoids} \cite{KaufmanRousseeuw1990PAM_k_medoids}.}
Initializing with the \texttt{k-center} solution, we utilize the PAM (Partitioning Around Medoids) refinement algorithm which swaps model embeddings under sum-similarity objective and replaces embeddings which result in improvement of the objective. 

Define the sum-similarity objective as 
\begin{equation} \label{eq:sum_similarity_objective}
J(\mathcal{M}_\textrm{sel}) \;=\; \sum_{i\in \mathcal{M}} \max_{s\in \mathcal{M}_\textrm{sel}} K(z_i,z_s),
\end{equation}
so that maximizing $J$ is equivalent to maximizing the total similarity coverage in \cref{eq:kmedoids_obj_kernel}.
PAM searches for swaps $s\in \mathcal{M}_\textrm{sel}$ and $h\notin \mathcal{M}_\textrm{sel}$ that improve the objective value:
\begin{equation}
\mathcal{M}_\textrm{sel}' \;=\; (\mathcal{M}_\textrm{sel}\setminus \{s\}) \cup \{h\},
\qquad \text{accept if } J(\mathcal{M}_\textrm{sel}') > J(\mathcal{M}_\textrm{sel}),
\end{equation}
and stops when no improving swap exists (or after a fixed number of iterations).

\paragraph{Random baseline.}
As a baseline, we sample a subset of models $\mathcal{M}_\textrm{sel}\subseteq \mathcal{M}$ of size $k$ uniformly at random, and report mean $\pm$ standard deviation in routing accuracy across trials.

\paragraph{Evaluation.}
For each $k$, we select $\mathcal{M}_\textrm{sel}$ using only $\{z_m\}$ (via the kernel $K(\cdot,\cdot)$) and evaluate $\mathrm{Acc}(\mathcal{M}_\textrm{sel})$ by restricting routing accuracy to the chosen subset of models.

\paragraph{Observations.} 
For the count constraint model portfolio selection problem, we find that simple approaches like \texttt{k-medoids} and \texttt{k-center} (\cref{fig:model_portfolio_joint}) achieve full model pool (comprising $112$ models) routing accuracy with just $15-20$ models in the selected subset, signifying   that selecting a subset of models by maximizing coverage over the model embedding space leads to capturing the collective abilities of the entire model pool. Such efficient selection of capable model subsets/portfolios points towards an efficient management of a large pool of language models, and eliminating redundancy in model abilities for simplifying deployment and improving hardware utilization. We reiterate that the presented approaches do not require any query-correctness prediction over models on representative queries, and rather the model subset choice is made purely based on model embedding space.

\subsubsection{Parameter Budget Constraint Selection}\label{subsubapp:parameter_constrained_portfolio_selection}

We utilize the sum-similarity objective as presented above in \cref{eq:sum_similarity_objective} and solve the parameter-budget-constrained maximization for respecting total parameter budget for model subset choice
\begin{equation}
\max_{\mathcal{M}_\mathrm{sel}\subseteq\mathcal{M}}\;\; J(\mathcal{M}_\mathrm{sel})
\quad \text{s.t.}\quad
\sum_{m\in \mathcal{M}_\mathrm{sel}} c_m \le \mathcal{C}.
\label{eq:knapsack_objective}
\end{equation}

\paragraph{Greedy selection (marginal gain per parameter).}
For simplicity, we study the greedy solution to the formulation in \cref{eq:knapsack_objective}. Let $b_i=\max_{s\in \mathcal{M}_\mathrm{sel}}K(z_i,z_s)$ denote most similar selected model in $\mathcal{M}_\textrm{sel}$ for model $i$.
For a candidate $m\notin \mathcal{M}_\mathrm{sel}$, the marginal gain is
\begin{equation}
\Delta(m\mid \mathcal{M}_\mathrm{sel})
\;=\;
J(\mathcal{M}_\mathrm{sel}\cup\{m\})-J(\mathcal{M}_\mathrm{sel})
\;=\;
\sum_{i\in\mathcal{M}} \max\big\{0,\;K(z_i,z_m)-b_i\big\}.
\label{eq:marginal_gain}
\end{equation}
Under the remaining budget, we add the model with the largest gain per parameter:
\begin{equation}
m^\star
\;=\;
\argmax_{\substack{m\notin \mathcal{M}_\mathrm{sel},\\\; c_m \leq~ (\mathcal{C}-\sum_{s\in \mathcal{M}_\mathrm{sel}} c_s)}}
\frac{\Delta(m\mid \mathcal{M}_\mathrm{sel})}{c_m},
\qquad
\mathcal{M}_\mathrm{sel} \leftarrow \mathcal{M}_\mathrm{sel} \cup \{m^\star\},
\label{eq:greedy_ratio}
\end{equation}
and stop when no feasible model remains.

\paragraph{Evaluation.}
For each parameter count constraint $\mathcal{C}$, we select $\mathcal{M}_\mathrm{sel}^\mathcal{C}$ using \cref{eq:greedy_ratio} and evaluate $\mathrm{Acc}(\mathcal{M}_\mathrm{sel}^\mathcal{C})$ by restricting evaluation to models present in $\mathcal{M}_\mathrm{sel}^\mathcal{C}$.
We also compare to a \emph{random feasible} baseline that samples subsets satisfying $\sum_{m\in S}c_m\leq \mathcal{C}$ and reports mean $\pm$ standard deviation across trials.

\paragraph{Observations.}
For the parameter constrained choice of models (\cref{fig:model_portfolio_joint}), we note that the described greedy algorithm which maximizes coverage over the space of model embeddings consistently outperforms random choice of models, attaining very close to full model pool routing accuracy at $\approx$150B parameters (full model pool has 112 models, with 1930B parameters in total). For practical deployment scenarios with limited hardware, clever choice of subset of models achieves almost full model pool's routing accuracy by eliminating duplicated model capabilities (in total just $8\%$ parameter budget across all models in the pool).

\section{Regenerating Model Embeddings for Backpropagation-based Approaches} \label{app:regenerating_embedllm_embeddings}

In this section, we describe the shortcomings of learnable embeddings based approaches. For the sake of brevity, we consider EmbedLLM \cite{ICLR2025_embedllm} and examine the geometry and capability representation of model embeddings learned by original training vs retraining/regenerating embeddings with the same data. Concretely, consider $\{e^{(m)}_\theta\}_{m\in\mathcal{M}}$ as the trainable embeddings for each model in the pool, and let $G_\psi$ denote the correctness predictor which is of the form $G_\psi (e^{(m)}_\theta, x) = \mathrm{MLP}_\psi(e^{(m)}_\theta \odot W_\psi x)$, where $x$ is the query encoding generated through a sentence encoder, $W_\psi$ is the trained query downprojection matrix, $\mathrm{MLP}_\psi$ is a multilayer perceptron network, and $\odot$ denotes elementwise product. We perform training over all model-query evaluation pairs, to minimize the binary cross entropy loss between predictions and observed correctness values from evaluations:
\begin{equation}\label{eq:embedllm_objective}
    \min_{\theta,\psi} \quad \mathbb{E}_m\;\mathbb{E}_x\;\; \Big[\mathrm{BCE}\big(G_\psi (e^{(m)}_\theta, x), \;y^{(m)}(x)\big)\Big]
\end{equation}

This results in learned model embeddings $\{e^{(m)}_\theta\}_{m\in\mathcal{M}}$, and correctness predictor $G_\psi$. To evaluate the stability of the generated model embeddings, we freeze the correctness predictor parameters $G_\psi$, and then retrain the model embeddings to obtain $\{e^{(m)}_{\theta_\mathrm{re}}\}_{m\in\mathcal{M}}$ with the same model query evaluation data. Mathematically, we perform:
\begin{equation}\label{eq:relearn_embedllm_objective}
    \min_{\theta_\mathrm{re}} \quad \mathbb{E}_m\;\mathbb{E}_x\;\; \Big[\mathrm{BCE}\big(G_\psi (e^{(m)}_{\theta_\mathrm{re}}, x), \;y^{(m)}(x)\big)\Big]
\end{equation}

This represents a practical scenario where unseen models are added to the pool, and embeddings need to be generated for them without affecting correctness predictor $G_\psi$ and already learned embeddings $\{e^{(m)}_\theta\}_{m\in\mathcal{M}}$ for models in the pool. For the sake of analysis, we consider new added models as clones of existing models with identical query evaluation data, and hence compare their newly generated embeddings $\{e^{(m)}_{\theta_\mathrm{re}}\}_{m\in\mathcal{M}}$ with originally trained embeddings $\{e^{(m)}_{\theta}\}_{m\in\mathcal{M}}$ for similarity.

Because the training objective in \cref{eq:relearn_embedllm_objective} differs from that in \cref{eq:embedllm_objective}, we anticipate small changes in the regenerated model embeddings, even though the embeddings are learned on the same query evaluation data. However, they should remain close to the original embeddings to preserve the practically desirable properties described in \cref{sec:introduction}.
If the regenerated embeddings $\{e^{(m)}_{\theta_\mathrm{re}}\}_{m\in\mathcal{M}}$  are very different as compared to original model embeddings $\{e^{(m)}_{\theta}\}_{m\in\mathcal{M}}$, we cannot utilize informative embedding geometry. In such scenarios, we cannot mutually compare model embeddings across models to identify similarities or differences in capabilities, making most of the analyses and practical applications presented in \cref{sec:experiments} infeasible. 

\begin{figure}[htbp]
  \centering
  \begin{subfigure}[b]{0.48\textwidth}
    \centering
    \includegraphics[width=\linewidth]{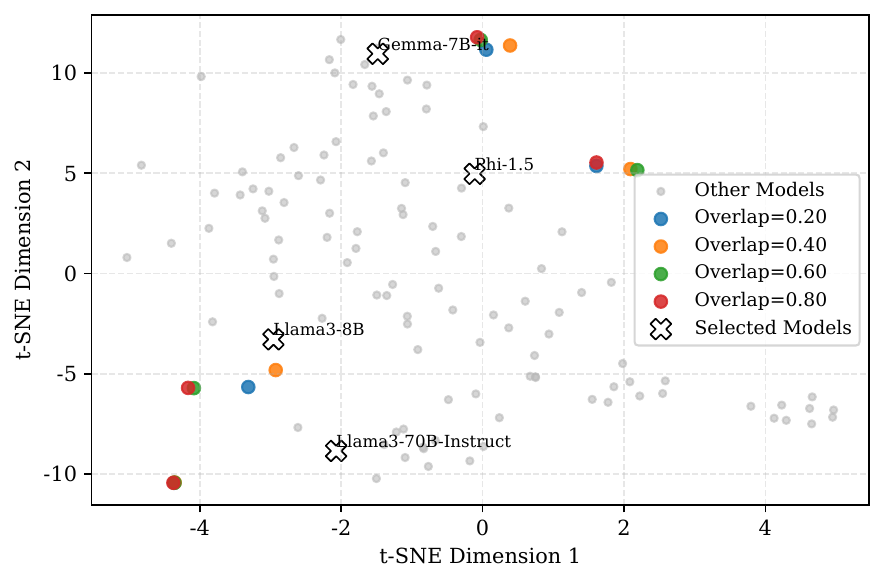}
    \caption{Regenerating model embeddings using query evaluations with varying overlap to original evaluation data.}
  \end{subfigure}
  \hfill
  \begin{subfigure}[b]{0.48\textwidth}
    \centering
    \includegraphics[width=\linewidth]{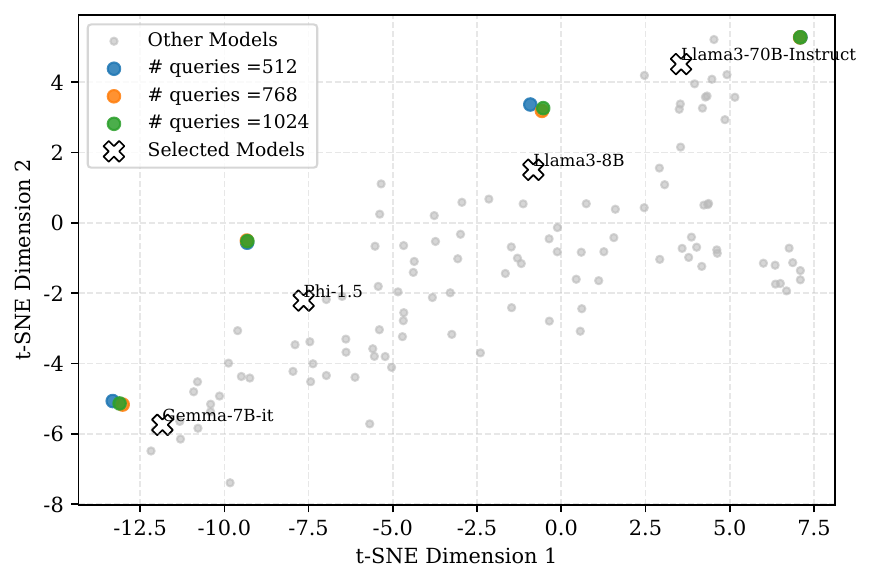}
    \caption{Regenerating model embeddings using queries sampled as a subset of original evaluation data.}
  \end{subfigure}
  \caption{Geometrically comparing original $\{e^{(m)}_{\theta}\}_{m\in\mathcal{M}}$  and regenerated embeddings $\{e^{(m)}_{\theta_\mathrm{re}}\}_{m\in\mathcal{M}}$ for varying overlap fraction with original evaluation set, and subsampling fewer queries from original evaluation set. Regenerated embeddings are far away from original model embeddings in both cases, denoting weakly informative embedding geometry.}
  \label{fig:relearning_embedllm_embeddings}
\end{figure}

\begin{figure}[htbp]
  \centering

  \begin{minipage}[t]{0.49\textwidth}
    \centering
    \includegraphics[width=0.95\linewidth]{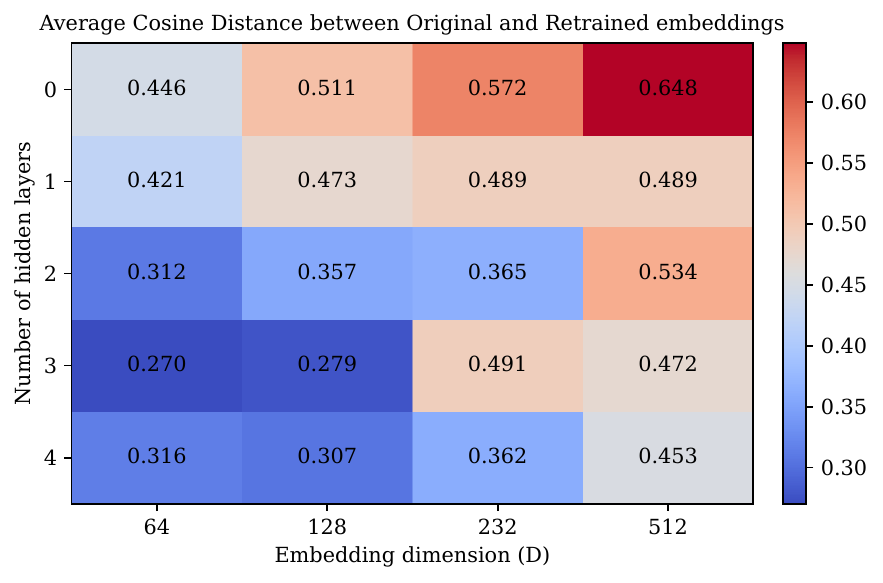}
    \captionof{figure}{Average cosine distance between initially trained model embeddings $\{e^{(m)}_{\theta}\}_{m\in\mathcal{M}}$ and regenerated model embeddings $\{e^{(m)}_{\theta_\mathrm{re}}\}_{m\in\mathcal{M}}$. We present results across number of hidden layers in $\mathrm{MLP}_\psi$ (with $d_\mathrm{hidden}=64$), along with chosen embedding dimension. Regenerated model embeddings exhibit high cosine distance to original trained embeddings even if retrained on same query evaluation data, rendering the model embedding geometry brittle for practical applications.}
    \label{fig:relear_embedllm_heatmap_cosine_distance}
  \end{minipage}\hfill
  \begin{minipage}[t]{0.49\textwidth}
    \centering
    \includegraphics[width=0.95\linewidth]{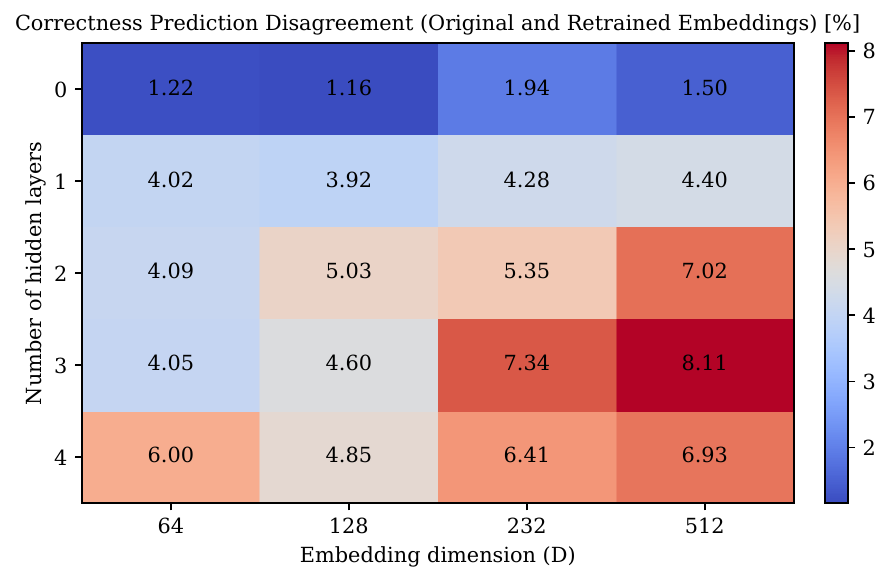}
    \captionof{figure}{Query Correctness Disagreement between initially trained model embeddings $\{e^{(m)}_{\theta}\}_{m\in\mathcal{M}}$ and regenerated model embeddings $\{e^{(m)}_{\theta_\mathrm{re}}\}_{m\in\mathcal{M}}$. We present the absolute percentage disagreement on correctness prediction averaged over all models for a common set of queries, across number of hidden layers in $\mathrm{MLP}_\psi$ (with $d_\mathrm{hidden}=64$), along with chosen embedding dimension. High correctness disagreement between original and regenerated embeddings denotes nonidentical capability representation, even if trained on the same query evaluation data.}
    \label{fig:relearn_embedllm_heatmap_test_disagreement}
  \end{minipage}

\end{figure}

We examine EmbedLLM \cite{ICLR2025_embedllm} through a similar approach as described in \cref{subsec:exp_embedding_stability_robustness}, where we regenerate model embeddings through new query evaluation sets to test sensitivity to \emph{which} queries are used for generating embeddings and \emph{how many} queries are used to generate embeddings. \Cref{fig:relearning_embedllm_embeddings} denotes geometric representation of original and regenerated embeddings (cf. \cref{subsec:exp_embedding_stability_robustness}), highlighting that regenerated embeddings are not in close proximity of the original embeddings. This is also presented numerically in \cref{fig:relear_embedllm_heatmap_cosine_distance}, where average cosine distance between respective original and regenerated embeddings is high, even though regenerated embeddings are trained on the same evaluation data as original embeddings. 

Discrepancy between original and regenerated embeddings also affects correctness prediction over queries. \Cref{fig:relearn_embedllm_heatmap_test_disagreement} denotes average correctness prediction disagreement (over a common set of test queries) by using original versus regenerated embeddings with the same correctness predictor $G_\psi$. Again, regenerated embeddings are trained on the original evaluation set, but still display up to $8.1\%$ disagreement rate on predicting correctness values for test queries.

\newpage

\section{Time Complexity of \AlgName{} for Generating Embeddings and Correctness Prediction} \label{app:time_complexity_locus}
In this section, we examine the time complexity of \AlgName{} for generating embeddings using query evaluations, and for correctness prediction on unseen queries. We utilize a single NVIDIA Tesla V100 (32GB) GPU \cite{nvidia-tesla-v100-datasheet-2018} for our experiments.

\begin{figure}[htbp]
  \centering
  \begin{subfigure}[b]{0.49\textwidth}
    \centering
    \includegraphics[width=0.8\linewidth]{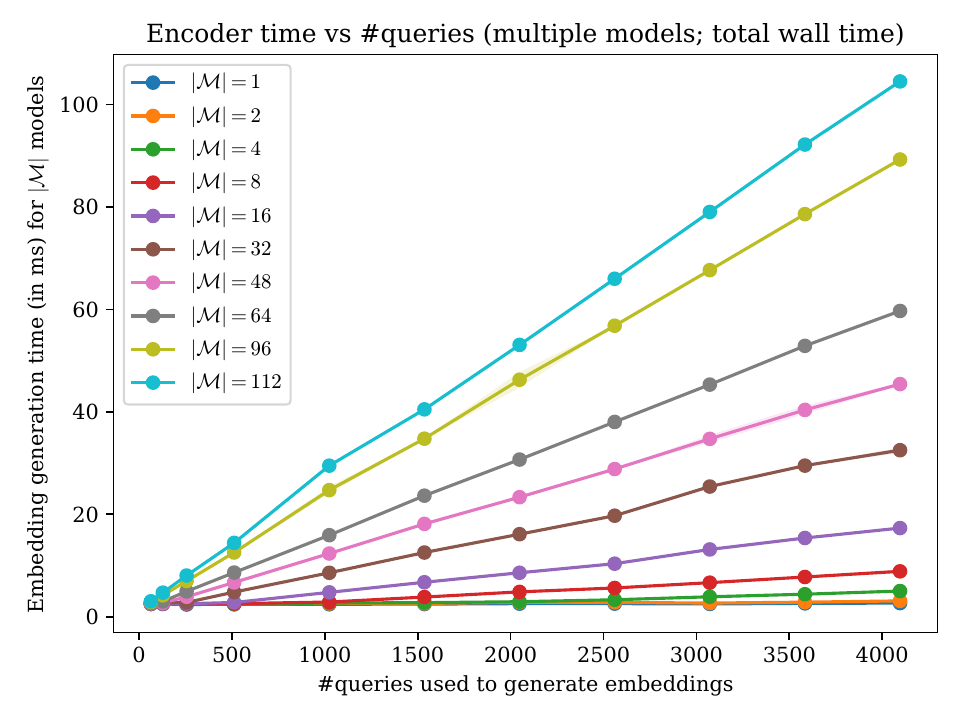}
    \caption{Model Embedding Generation (via $F_\theta$)}
    \label{subfig:time_complexity_locus_encoder}
  \end{subfigure}\hfill
  \begin{subfigure}[b]{0.49\textwidth}
    \centering
    \includegraphics[width=0.8\linewidth]{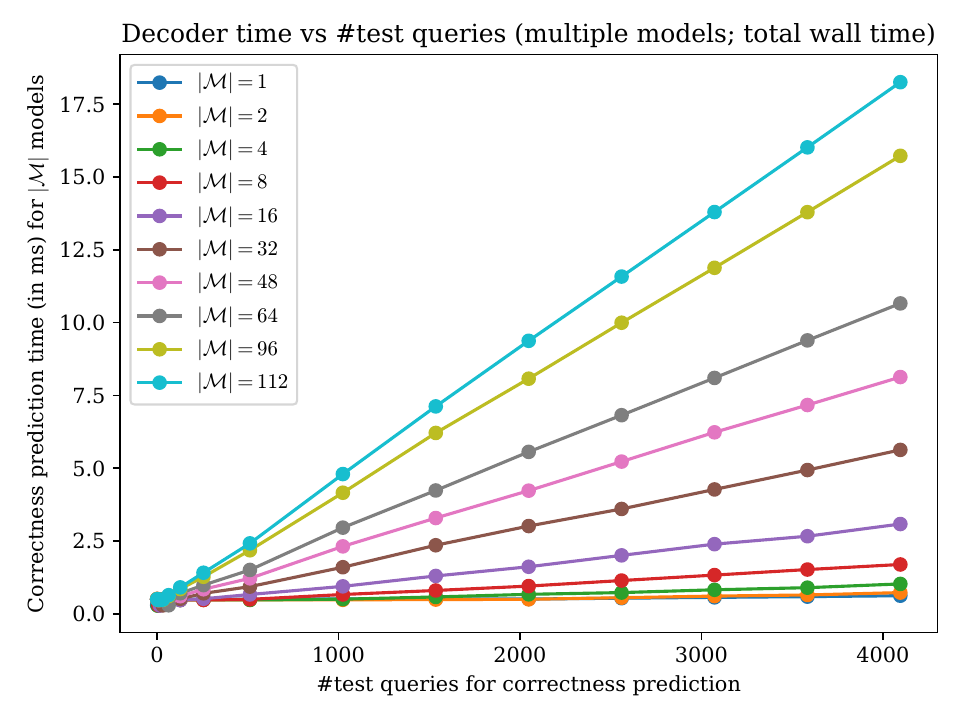}
    \caption{Unseen Query correctness prediction (via $G_\psi$)}
    \label{subfig:time_complexity_locus_decoder}
  \end{subfigure}
  \caption{\textbf{Time Complexity of \AlgName{} (in ms).} \textbf{\textit{(a)}} We consider embedding generation for different number of models ($|\mathcal{M}|$), using varying query evaluation sets. Linear time complexity (in terms of \# evaluation queries used) of our Latent Bottleneck Attention blocks (\cref{eq:latent_bottleneck_tblock,subsec:methods_encoder}) is evident from above. \textbf{\textit{(b)}} The time complexity of our correctness predictor is of the order of $\approx$20ms, even for calculating probability of correctness over the entire model pool (with 112 models) for a batch of 4096 unseen queries. This denotes the efficiency of our correctness predictor for practical tasks such as routing, where the additional correctness prediction overhead is orders of magnitude smaller than the actual response generation time (usually in seconds \cite{10.1145/3600006.3613165_vllm}).}
  \label{fig:time_complexity_locus}
\end{figure}

\Cref{subfig:time_complexity_locus_encoder} denotes the wall clock time required to generate model embeddings (for $|\mathcal{M}|$ models in parallel) using query evaluation sets of varying sizes. We observe that even for parallel embedding generation for the full model pool (112 models) on 4096 queries, required time is $\approx$100ms. This indicates powerful yet efficient design of \AlgName{} for generating model embeddings from observed query evaluations. Additionally, \cref{subfig:time_complexity_locus_decoder} denotes the correctness prediction time for varying number of unseen queries using the decoder $G_\psi$. Calculating correctness probabilities over the entire model pool for 4096 queries requires $\approx$20ms, which is orders of magnitude smaller than actual response generation time from language models (usually in seconds \cite{10.1145/3600006.3613165_vllm}). This further strengthens the practicality of \AlgName{} for applications like routing, model ranking etc.

\section{Implementation Details and Reproducibility Checklist} \label{app:implementation_details_reproducibility_checklist}
We have provided training code and data processing scripts for \AlgName{} in the attached link \href{https://github.com/patel-shivam/locus_code_release}{https://github.com/patel-shivam/locus\_code\_release}. Across experiments in this work, \AlgName{} utilizes $r=64$ learnable vectors for Latent bottleneck Attention Blocks (\cref{eq:latent_bottleneck_tblock}), and our default model embedding dimension is $d=128$. We choose query, key and value dimensions same as $d$ throughout all the attention layers. The feedforward network ($\mathrm{FFN}$) is a two-layer \textrm{MLP} with hidden dimension $d_\mathrm{ff} = 2d$. The tokenizer module $h_\omega$ is a single layer \textrm{MLP} with output dimension as $d$. Correctness predictor $G_\psi$ is a 2-layer \textrm{MLP} with hidden dimension 64.

For training encoder-decoder pair ($F_\theta,G_\psi$) in \cref{subsec:methods_training}, we sample the query evaluations as input to the encoder $S_m^\mathrm{enc}\subseteq S_m$ to generate model embeddings $z_m$, and independently sample a subset $S_m^\mathrm{dec}\subseteq S_m$ for training the correctness predictor ($S_m^\mathrm{enc}$ and $S_m^\mathrm{dec}$ can have overlap, but only during training). The number of queries chosen in $(S_m^\textrm{enc}, S_m^\textrm{dec})$ can be varied to adjust to training hardware, our choices are detailed in the supplementary code submission. For all of the presented numerical experiments on correctness prediction and routing accuracy, test queries are only used for evaluating the decoder performance $G_\psi$, and never used for generating the embeddings $\{z_m\}_{m\in\mathcal{M}}$. This ensures that there is no leakage of test data into generating the model embeddings. Additionally, experiments examining correctness agreement rate utilize test set of common queries over which model correctness agreement is evaluated (\cref{fig:performance_distance_correlation,fig:nearest_neighbor_proxies}), again ensuring no test queries used for generating embeddings and no training queries used for evaluation.

%%%%%%%%%%%%%%%%%%%%%%%%%%%%%%%%%%%%%%%%%%%%%%%%%%%%%%%%%%%%%%%%%%%%%%%%%%%%%%%
%%%%%%%%%%%%%%%%%%%%%%%%%%%%%%%%%%%%%%%%%%%%%%%%%%%%%%%%%%%%%%%%%%%%%%%%%%%%%%%

\end{document}